\documentclass{article}

    \PassOptionsToPackage{numbers, compress}{natbib}

\usepackage[preprint]{neurips_2025}

\usepackage[utf8]{inputenc} 
\usepackage[T1]{fontenc}    
\usepackage{hyperref}       
\usepackage{url}            
\usepackage{booktabs}       
\usepackage{amsfonts}       
\usepackage{nicefrac}       
\usepackage{microtype}      
\usepackage{xcolor}         
\usepackage{xspace}

\usepackage{enumitem}
\usepackage{multirow}
\usepackage[figuresright]{rotating}
\usepackage{makecell}
\usepackage{subcaption}

\usepackage{subcaption}
\usepackage{pifont}
\usepackage{environ}

\definecolor{GeoGround}{RGB}{255,0,0}
\definecolor{VHM}{RGB}{0,0,255}
\definecolor{Falcon}{RGB}{255,165,0}
\definecolor{GeoChatZero}{RGB}{128,0,128}

\usepackage{dirtree}
\usepackage{amsmath}

\usepackage{fontawesome5}
\definecolor{forestgreen}{rgb}{0.0, 0.5, 0.0}
\definecolor{ashgrey}{rgb}{0.7, 0.75, 0.71}
\definecolor{darkorange}{rgb}{1.0, 0.55, 0.0}
\usepackage{colortbl}
\definecolor{LightOrange}{rgb}{1,0.95,0.88}
\newcommand{\icoyes}{\textcolor{forestgreen}{\faIcon{check-circle}}\xspace}

\newcommand{\icono}{\textcolor{ashgrey}{\faIcon{times-circle}}\xspace}

\newcommand{\papernameAbbrev}{REOBench}

\newcommand{\gitpage}{https://github.com/lx709/REOBench}

\newcommand{\huggingpage}{https://huggingface.co/datasets/xiang709/REOBench}

\usepackage{graphicx}
\graphicspath{{AID_noise/}{figs_diss/}}

\title{\papernameAbbrev{}: Benchmarking Robustness of Earth Observation Foundation Models}

\author{%
\bfseries
  Xiang Li$^{1}\thanks{First three authors contributed equally.}$ ,
  Yong Tao$^2 \footnotemark[1]$ ,
  Siyuan Zhang$^3\footnotemark[1]$ ,
  Siwei Liu$^4$,
  Zhitong Xiong$^5$  \\
  \textbf{Chunbo Luo}$^2$, 
  \textbf{Lu Liu}$^2$, 
  \textbf{Mykola Pechenizkiy}$^6$,
  \textbf{Xiao Xiang Zhu}$^5$,
  \textbf{Tianjin Huang}$^{2,6} \thanks{Corresponding author: \texttt{t.huang2@exeter.ac.uk}}$ \\
  \\
  $^1$ University of Bristol, UK 
  $^2$ University of Exeter, UK \\
  $^3$ South China Normal University, China 
  $^4$ The University of Aberdeen, UK \\
  $^5$ Technical University of Munich, Germany 
  $^6$ Eindhoven University of Technology, NL 
}

\begin{document}

\maketitle

\begin{abstract}
Earth observation foundation models have shown strong generalization across multiple Earth observation tasks, but their robustness under real-world perturbations remains underexplored. To bridge this gap, we introduce \papernameAbbrev{}, the first comprehensive benchmark for evaluating the robustness of Earth observation foundation models across six tasks and twelve types of image corruptions, including both appearance-based and geometric perturbations. To ensure realistic and fine-grained evaluation, our benchmark focuses on high-resolution optical remote sensing images, which are widely used in critical applications such as urban planning and disaster response. We conduct a systematic evaluation of a broad range of models trained using masked image modeling, contrastive learning, and vision-language pre-training paradigms. 
Our results reveal that \ding{182} existing Earth observation foundation models experience significant performance degradation when exposed to input corruptions. \ding{183} The severity of degradation varies across tasks, model architectures, backbone sizes, and types of corruption, with performance drop varying from less than 1\% to over 25\%. \ding{184} Vision-language models show enhanced robustness, particularly in multimodal tasks. \papernameAbbrev{} underscores the vulnerability of current Earth observation foundation models to real-world corruptions and provides actionable insights for developing more robust and reliable models. Code and data are publicly available at \gitpage.


\end{abstract}

\section{Introduction}

Recent studies have shown that foundation models pre-trained on large-scale datasets have demonstrated powerful capabilities across multiple domains. Models such as MAE~\cite{mae}, CLIP~\cite{clip}, MiniGPT-4~\cite{minigpt4}, and LLaVA~\cite{llava} have achieved remarkable success in multiple vision and vision-language tasks. These models are capable of extracting representative features from images, and more importantly, can be quickly adapted to multiple downstream tasks with minimal fine-tuning, significantly improving the efficiency and effectiveness of task-solving. 

In the field of remote sensing, the rapid growth in data volume has sparked significant interest in developing foundation models for remote sensing image analysis. Earth observation foundation models (EOFMs) aim to leverage supervised or self-supervised training to build large-scale pre-trained models that can be adapted to a wide range of downstream tasks, thereby improving the performance and efficiency of Earth observation applications. In recent years, a growing body of research has focused on constructing such foundation models tailored for remote sensing. Mainstream models can be divided into unimodal pre-trained foundation models (e.g., SatMAE~\cite{satmae}, RingMo~\cite{ringmo}, ScaleMAE~\cite{scalemae}, SpectralGPT~\cite{hong2024spectralgpt}) and vision-language foundation models (e.g., RemoteCLIP~\cite{remoteclip}, GeoRSCLIP~\cite{rs5m}, RSGPT~\cite{rsgpt}, GeoChat~\cite{geochat}). These EOFMs have demonstrated their powerful performance in numerous downstream tasks. A comprehensive review of EOFMs can be found in~\cite{jiao2023brain,li2024vision,zhu2024foundations,xiao2024foundation,lu2025vision,huo2025remote,huang2025survey}.

Despite advancements in EOFMs, there remains a significant gap in systematically benchmarking their robustness towards image perturbations. Remote sensing images are particularly susceptible to factors such as weather conditions or sensor discrepancies, which can introduce significant noise and variability~\cite{Kazmi2023AdversarialAO,Mei2023ACS,lian2022benchmarking}, posing challenges to current EOFMs. Therefore, developing a comprehensive benchmark to evaluate and compare the robustness of these models holds great academic and practical significance. Such benchmarking efforts can guide the design of highly robust EOFMs that effectively adapt to noise and data variability, ensuring stable and reliable results under diverse conditions.

To achieve this, we introduce \textbf{\papernameAbbrev}, a comprehensive \textbf{Bench}mark designed to evaluate the \textbf{R}obustness of \textbf{E}arth \textbf{O}bservation foundation models, covering state-of-the-art models based on masked image modeling, contrastive learning, and large language models. \papernameAbbrev{} focuses on high-resolution optical remote sensing images, which are widely used in real-world applications such as urban planning and disaster response. We conducted experiments on \textbf{six} widely studied remote sensing image understanding tasks, covering both vision-centric and vision-language tasks, under \textbf{twelve} types of perturbations. These include both appearance-based corruptions (e.g., noise, blur, haze) and geometric distortions (e.g., rotation, scale, translation), applied at varying severity levels to simulate realistic environmental and sensor-induced challenges. Our evaluation yields three key findings:
\begin{itemize}[leftmargin=*]
    \item[$\star$] Existing Earth observation foundation models suffer noticeable performance degradation under common image corruptions, with particularly sharp drops for the models based on masked image modeling.
    \item[$\star$] The degree of vulnerability to image corruptions varies across tasks, model architectures, and types of perturbations, with performance drop varying from less than 1\% to over 20\%.
    \item[$\star$] Vision-language foundation models exhibit greater robustness to visual perturbations compared to vision-centric foundation models, particularly in image-level scene classification tasks.
\end{itemize}

In summary, \papernameAbbrev{} provides the first large-scale, task-diverse, and perturbation-rich benchmark for evaluating robustness in EOFMs. It offers actionable insights for the research community and serves as a stepping stone toward building more reliable, generalizable, and trustworthy AI systems for Earth observation.

\section{\papernameAbbrev{} Dataset}

To systematically evaluate the robustness of EOFMs, we construct a benchmark dataset by incorporating widely used remote sensing datasets spanning diverse tasks. Specifically, we include AID~\cite{xia2017aid} for scene classification, ISPRS Potsdam~\cite{rottensteiner2012isprs} for semantic segmentation, DIOR~\cite{dior} for object detection, and three subsets from VRSBench~\cite{vrsbench} for image captioning, visual question answering (VQA), and visual grounding. These datasets are selected based on their popularity, diversity of content, and relevance to the tasks under evaluation.


\subsection{Corruptions in Remote Sensing Images}

Remote sensing platforms are subject to a wide range of visual degradations that differ significantly from those encountered by ground-based cameras. To systematically evaluate the robustness of  RSFMs, we construct a benchmark comprising \textbf{12 synthetic corruptions}, categorized into three types: \emph{environmental}, \emph{sensor-induced}, and \emph{geometric}. Each corruption is generated using physically or statistically grounded procedures to ensure the resulting images remain photorealistic while faithfully reflecting failure modes commonly observed in satellite and UAV imagery.

\textbf{Environmental Corruptions.}
Atmospheric and illumination variations constitute predominant environmental degradations. For instance, \emph{Cloud} occlusions substantially obscure optical remote sensing data, severely impacting scene interpretability \cite{jeppesen2019cloud, sarukkai2020cloud}. Variations in \emph{Brightness} resulting from shifting sun angles affect radiometric stability and degrade feature matching and object recognition performance \cite{zhang2016deep, muller2019brightness}. \emph{Haze}, caused by aerosol scattering, significantly lowers image contrast and impairs detection and classification accuracy \cite{makarau2014haze, li2025dhc}. Following established benchmarks \cite{hendrycks2019benchmarking}, we simulate these environmental corruptions using physically motivated image augmentation techniques.

\textbf{Sensor-induced Corruptions.}
Imperfections during sensor capture or data transmission introduce various degradations. \emph{Gaussian Blur}, indicative of defocusing or modulation transfer function (MTF) degradation, compromises tie-point accuracy and feature localization \cite{sieberth2015uav, sieberth2014influence}. \emph{Motion Blur}, arising from platform vibrations or rapid movements, negatively impacts object detection and tracking in aerial inspections \cite{sieberth2014motion}. \emph{Gaussian Noise} and \emph{Salt \& Pepper Noise}, simulating electronic interference and bit-flip errors respectively, significantly decrease segmentation and classification accuracy \cite{boonprong2018classification, narayanan2003effects}. \emph{Sensor Gap} degradations, exemplified by the Landsat-7 SLC-off issue, necessitate specialized gap-filling methodologies \cite{chen2011simple, chen2012making}. Furthermore, \emph{Compression} artifacts, such as those from JPEG/JPEG2000, substantially impair the quality of CNN feature extraction \cite{benbarrad2021impact}. These sensor-induced corruptions are replicated through established augmentation and simulation protocols in line with existing research \cite{hendrycks2019benchmarking, dong2023benchmarking}.

\textbf{Geometric Corruptions.}
Geometric distortions originate primarily from variations in sensor orientation, altitude, and registration accuracy. \emph{Rotation} caused by platform roll or yaw introduces inconsistencies in orientation-sensitive feature extraction processes \cite{kang2021rotation}. \emph{Scale} alterations resulting from altitude fluctuations pose significant challenges for detectors lacking robust multi-scale adaptability \cite{han2022context}. \emph{Translation}, modeling inaccuracies due to GPS drift, registration errors, or parallax, adversely affects pixel-aligned or patch-based analysis methods \cite{lee2021cnn}. To effectively simulate these geometric degradations, we apply spatial transformations, including image \emph{Rotation}, \emph{Scaling}, and \emph{Translation}, to remote sensing images.

\begin{figure*}[ht!]
    \centering

    \begin{subfigure}[t]{0.16\textwidth}
        \centering
        \includegraphics[width=\textwidth]{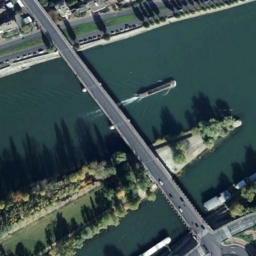}
        \caption{Original}
    \end{subfigure}
    \begin{subfigure}[t]{0.16\textwidth}
        \centering
        \includegraphics[width=\textwidth]{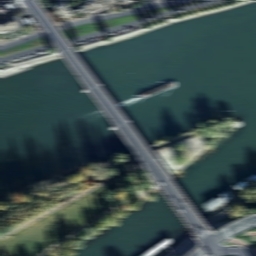}
        \caption{Motion blur-1}
    \end{subfigure}
    \begin{subfigure}[t]{0.16\textwidth}
        \centering
        \includegraphics[width=\textwidth]{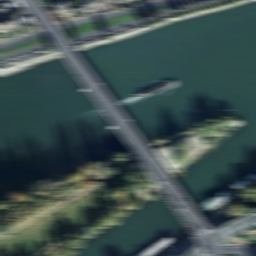}
        \caption{Motion blur-2}
    \end{subfigure}
    \begin{subfigure}[t]{0.16\textwidth}
        \centering
        \includegraphics[width=\textwidth]{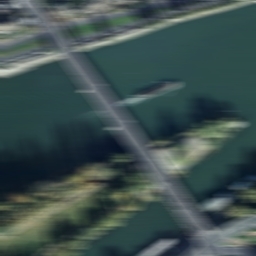}
        \caption{Motion blur-3}
    \end{subfigure}
    \begin{subfigure}[t]{0.16\textwidth}
        \centering
        \includegraphics[width=\textwidth]{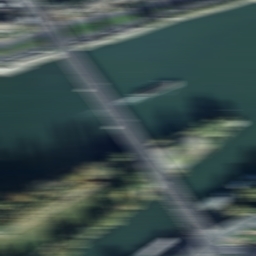}
        \caption{Motion blur-4}
    \end{subfigure}
    \begin{subfigure}[t]{0.16\textwidth}
        \centering
        \includegraphics[width=\textwidth]{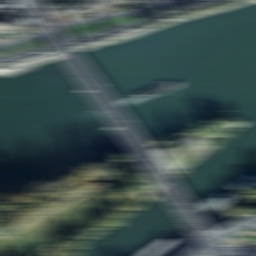}
        \caption{Motion blur-5}
    \end{subfigure}

    \begin{subfigure}[t]{0.16\textwidth}
        \centering
        \includegraphics[width=\textwidth]{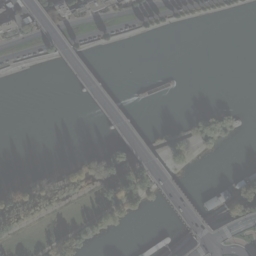}
        \caption{Brightness}
    \end{subfigure}
    \begin{subfigure}[t]{0.16\textwidth}
        \centering
        \includegraphics[width=\textwidth]{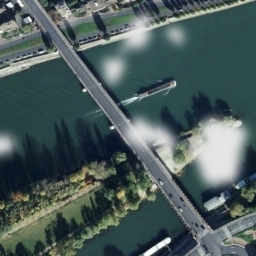}
        \caption{Cloud}
    \end{subfigure}
    \begin{subfigure}[t]{0.16\textwidth}
        \centering
        \includegraphics[width=\textwidth]{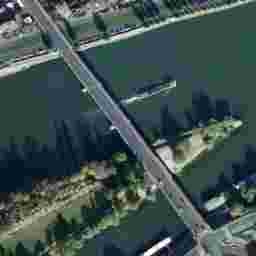}
        \caption{Compression}
    \end{subfigure}
    \begin{subfigure}[t]{0.16\textwidth}
        \centering
        \includegraphics[width=\textwidth]{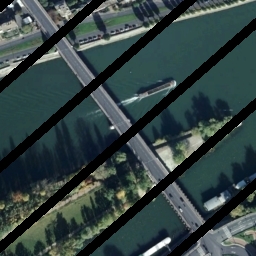}
        \caption{Gap}
    \end{subfigure}
    \begin{subfigure}[t]{0.16\textwidth}
        \centering
        \includegraphics[width=\textwidth]{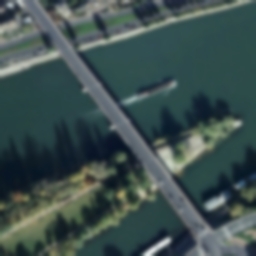}
        \caption{Gauss Blur}
    \end{subfigure}
    \begin{subfigure}[t]{0.16\textwidth}
        \centering
        \includegraphics[width=\textwidth]{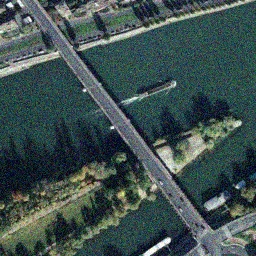}
        \caption{Gauss Noise}
    \end{subfigure}

    \begin{subfigure}[t]{0.16\textwidth}
        \centering
        \includegraphics[width=\textwidth]{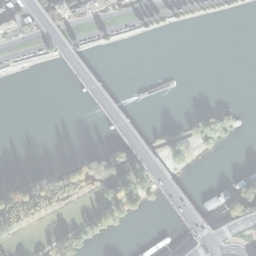}
        \caption{Haze}
    \end{subfigure}
    \begin{subfigure}[t]{0.16\textwidth}
        \centering
        \includegraphics[width=\textwidth]{AID_noise/motion_blur_severity_5_bridge_231.jpg}
        \caption{Motion Blur}
    \end{subfigure}
    \begin{subfigure}[t]{0.16\textwidth}
        \centering
        \includegraphics[width=\textwidth]{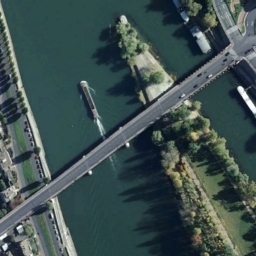}
        \caption{Rotate}
    \end{subfigure}
    \begin{subfigure}[t]{0.16\textwidth}
        \centering
        \includegraphics[width=\textwidth]{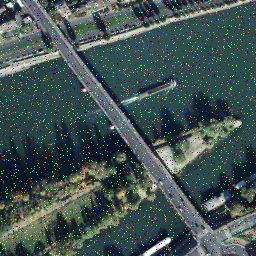}
        \caption{Salt \& Pepper}
    \end{subfigure}
    \begin{subfigure}[t]{0.16\textwidth}
        \centering
        \includegraphics[width=\textwidth]{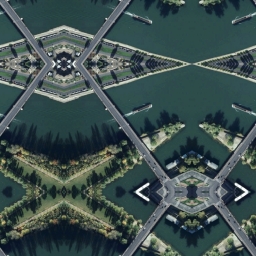}
        \caption{Scale}
    \end{subfigure}
    \begin{subfigure}[t]{0.16\textwidth}
        \centering
        \includegraphics[width=\textwidth]{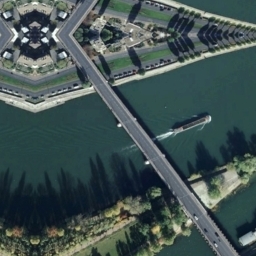}
        \caption{Translate}
    \end{subfigure}
    
    \label{fig:noisy}
    \caption{Example of perturbed images. In the first row, we present the original clean image alongside images perturbed by five levels of motion blur. The second and third rows illustrate examples of images corrupted by a range of perturbation types.}
    \label{fig:clean_noisy}
\end{figure*}

In total, these corruption categories encompass twelve distinct types. Each type of corruption is applied consistently across all datasets at five severity levels. Fig.~\ref{fig:clean_noisy} illustrates one example of original and corrupted images.

\subsection{Definition of Corruption Robustness }

We formalize the corruption robustness of EOFMs as its ability to maintain task performance in the presence of realistic geophysical and sensor-induced degradations frequently encountered in Earth observation. Let
\(
f:\mathcal{X} \to \mathcal{Y}
\)
denote an EOFMs that maps an input image \(x \in \mathcal{X}\) to a label  \(y \in \mathcal{Y}\), with \((x, y)\) sampled from an underlying geospatial data-generating distribution \(\mathcal{D}\). We define a set of corruptions \(\mathcal{C} = \{c_1, \dots, c_K\}\), where each \(c_k \in \mathcal{C}\) represents a physically plausible corruption operator, such as haze, cloud occlusion, or sensor noise. Each corruption occurs with a non-zero prevalence \(\mathbb{P}_{\mathcal{C}}(c_k) > 0\). 
To quantify robustness, we define the \emph{relative task performance drop} (\(\mathcal{R}_{\text{TP}}\)), which measures the degradation in model performance under corrupted inputs relative to its accuracy on clean data:
\begin{equation}
\mathcal{R}_{\text{TP}}
\;=\;
\frac{
\mathbb{P}_{(x,y)\sim\mathcal{D}}\!\left[f(x)=y\right]
-
\mathbb{E}_{c\sim\mathcal{C}}\!\left[\mathbb{P}_{(x,y)\sim\mathcal{D}}\!\left[f(c(x))=y\right]\right]
}{
\mathbb{P}_{(x,y)\sim\mathcal{D}}\!\left[f(x)=y\right]
}.
\end{equation}
A \emph{smaller} \(\mathcal{R}_{\text{TP}}\) indicates greater robustness, as it reflects less relative degradation when encountering corrupted data.

\section{Benchmark Robustness on EOFMs}
We evaluate the robustness of EOFMs across six widely studied remote sensing image understanding tasks: scene classification, semantic segmentation, object detection, image captioning, visual question answering (VQA), and visual grounding. The evaluated models represent the current state-of-the-art in remote sensing and can be broadly categorized into the following three types.

\noindent \textbf{MIM-based foundation models.} Masked image modeling (MIM) has gained popularity through the pioneering work of MAE~\cite{mae}. In the field of remote sensing, notable approaches include SatMAE~\cite{satmae}, RVSA~\cite{rvsa}, ScaleMAE~\cite{scalemae}, and SatMAE++~\cite{satmae++}. 

\noindent \textbf{CL-based foundation models.} Building on the success of the pioneering work CLIP~\cite{clip}, multiple contrastive learning (CL) -based foundation models have been introduced in the field of remote sensing, such as RemoteCLIP~\cite{remoteclip} and GeoRSCLIP~\cite{rs5m}. To investigate robustness with respect to different backbone sizes, we evaluate two commonly used architectures in our experiments: ViT-B/32 and ViT-L/14~\cite{vit}.

\noindent \textbf{LLM-based foundation models}. Following the pioneering works of GPT-4~\cite{gpt4}, MiniGPT-4~\cite{minigpt4}, and LLaVA~\cite{llava}, multimodal large language models (MLLMs) have attracted significant research attention in recent years. Notable approaches include GeoChat~\cite{geochat}, LHRS-Bot~\cite{lhrs}, RS-LLaVA~\cite{rsllava}, VHM~\cite{vhm}, SkySenseGPT~\cite{skysensegpt}, and Falcon~\cite{falcon}. In our experiments, we evaluate models with open-access code and pretrained weights for comparison.

\subsection{Implementation Details}

For MIM- and CLIP-based models, we take the vision backbones from pretrained foundation models and append a task-specific head (e.g., MLP, detectors, or segmentors) for each task. For these LLM-based models, since these generalist models usually freeze their vision backbones and can naturally handle multiple tasks, we directly evaluate \textit{zero-shot} performance of these models to test their robustness.

For the scene classification task, we take the backbone from all pretrained foundation models and append a single linear layer after the backbone for classification. For the semantic segmentation (resp. detection) task, we follow RVSA~\cite{rvsa} to use UpperNet~\cite{uppernet} (resp. Oriented R-CNN~\cite{orcnn}) as the segmentor (resp. detector) and replace its backbone with that from pretrained foundation models. Following RVSA~\cite{rvsa}, we build a feature pyramid from blocks 4, 6, 8, and 12 using up/down-sampling. All models are trained for 12 epochs with an initial learning rate of 1e-5, decayed by 0.1 at epochs 8 and 11.

For vision-language tasks, including image captioning, visual question answering, and visual grounding, we follow the original paper designs to craft task-specific prompts and evaluate the zero-shot performance of these foundation models to assess robustness. It should be noted that different pretrained foundation models are designed to accept images at specific resolutions. When the input image size differs from the pretrained backbone's expected resolution, we interpolate the position embeddings in the backbone to accommodate the new input dimensions.

\subsection{Scene Classification}

From Table \ref{tab_cls}, we can draw the following findings: 1) All benchmark methods suffer from serious performance under image corruptions for the scene classification tasks, especially for MIM-based methods. 2) CL- and LLM-based methods are more robust towards image corruptions than MIM-based methods. This is probably because CL- and LLM-based methods are trained by matching image-text pairs in a shared embedding space, learning high-level semantic features less sensitive to low-level corruptions. In contrast, MIM-based methods are trained by reconstructing pixel- or token-level details, making them sensitive to local corruptions. Specifically, VHM~\cite{vhm} achieves the least performance drop under image corruptions. 3) CL-based methods usually perform better than MIM and LLM-based methods on the scene classification task, for both clean and noisy images. Specifically, GeoRSCLIP~\cite{rs5m} achieves the best scene classification performance under image corruptions.

\vspace{-4pt}
\begin{table}[h]
\centering
\caption{Scene classification performance on AID dataset across different image perturbations. ${zs}$ denotes zero-shot evaluation.}
\small 
\setlength{\tabcolsep}{4pt} 
\resizebox{\textwidth}{!}{
\begin{tabular}{@{}lcccccccccccccccc@{}}
\toprule
\multirow{2}{*}{Method}  & \multirow{2}{*}{Backbone} & \multirow{2}{*}{Clean} & Brightness & \multirow{2}{*}{Cloud} & Compression & Data & Gauss & Gauss & \multirow{2}{*}{Haze} & Motion & \multirow{2}{*}{Rotate} & Salt & \multirow{2}{*}{Scale} & \multirow{2}{*}{Translate} & \multirow{2}{*}{Avg} & \multirow{2}{*}{\(\mathcal{R}_{\text{TP}}\)} \\
 & & & Contrast & & Artifacts & Gaps & Blur & Noise &  & Blur &  & Pepper &  &  &  & \\
\hline
\multicolumn{17}{c}{MIM-based} \\
\hline
SATLAS~\cite{bastani2023satlaspretrain}  & Swin-B & 90.85 & 82.54 & 84.32 & 73.36 & 67.23 & 78.10 & 79.16 & 80.46 & 32.44 & 72.54 & 77.56 & 72.54 & 88.54 & 74.07 & {18.47}\\
SatMAE~\cite{satmae}  & ViT-L & 72.05 & 44.82 & 59.58 & 67.26 & 46.49 & 71.33 & 71.25 & 28.31 & 63.85 & 69.15 & 70.45 & 59.74 & 66.12 & 59.86 & {16.92}\\
Scale-MAE~\cite{scalemae}  & ViT-L & 75.75 & 51.80 & 72.65 & 39.60 & 43.69 & 31.65 & 46.31 & 55.24 & 17.49 & 66.15 & 47.27 & 61.58 & 69.84 & 50.27 & {33.64}\\
RVSA~\cite{rvsa}  & ViT-B & 84.60 & 56.84 & 77.33 & 56.07 & 53.14 & 53.53 & 32.51 & 49.19 & 23.45 & 76.88 & 35.12 & 71.78 & 77.22 & 55.26 & {34.69}\\
SatMAE++~\cite{satmae++}  & ViT-L & 91.35 & 64.62 & 82.64 & 62.69 & 60.70 & 48.23 & 76.98 & 62.56 & 29.43 & 85.49 & 73.22 & 75.79 & 87.61 & 67.50 & {26.11}\\
\hline
\multicolumn{17}{c}{CL-based} \\
\hline
RemoteCLIP$_{zs}$~\cite{remoteclip}  & ViT-L & 81.10 & 78.32 & 80.64 & 73.91 & 79.43 & 76.83 & 76.72 & 80.10 & 57.02 & 82.80 & 70.90 & 68.39 & 80.72 & 75.48 & {6.93}\\
RemoteCLIP~\cite{remoteclip}  & ViT-B & 96.85 & 90.80 & 95.36 & 91.13 & 88.96 & 89.18 & 94.25 & 87.46 & 63.75 & 96.22 & 91.43 & 83.62 & \textbf{95.42} & 88.97 & {8.15}\\
RemoteCLIP~\cite{remoteclip}  & ViT-L & 95.45 & 93.11 & 93.80 & 88.77 & 94.21 & 92.47 & 94.20 & 93.37 & \textbf{74.45} & 95.01 & 86.99 & 83.37 & 94.06 & 90.32 & {5.38}\\
GeoRSCLIP$_{zs}$~\cite{rs5m}  & ViT-L & 66.05 & 62.41 & 65.47 & 60.45 & 64.41 & 62.03 & 62.32 & 62.24 & 44.20 & 65.52 & 58.88 & 52.59 & 64.25 & 60.40 & {8.55}\\
GeoRSCLIP~\cite{rs5m}  & ViT-B & 96.90 & 93.59 & 96.04 & 91.01 & 93.34 & 92.60 & 92.99 & 92.91 & 57.78 & 95.70 & 88.22 & 75.70 & 93.87 & 88.65 & {8.51}\\
GeoRSCLIP~\cite{rs5m}  & ViT-L & \textbf{97.40} & \textbf{96.27} & \textbf{96.45} & \textbf{92.28} & \textbf{95.62} & \textbf{96.09} & \textbf{95.68} & \textbf{96.00} & 71.03 & \textbf{97.20} & \textbf{92.75} & \textbf{77.16} & 95.15 & \textbf{91.80} & {5.74}\\
\hline
\multicolumn{17}{c}{LLM-based} \\
\hline
GeoChat~\cite{geochat}  & ViT-L & 65.85 & 64.67 & 65.26 & 60.71 & 64.61 & 63.32 & 62.34 & 64.54 & 48.68 & 65.05 & 62.32 & 56.21 & 62.91 & 61.72 & {6.27}\\
LHRS-Bot~\cite{lhrs}  & ViT-L & 87.75 & 87.38 & 86.82 & 78.53 & 85.95 & 84.94 & 82.62 & 87.62 & 67.76 & 87.31 & 76.73 & 79.07 & 86.71 & 82.79 & {5.65}\\
RS-LLaVA~\cite{rsllava} & ViT-L & 67.55 & 65.36 & 68.95 & 63.05 & 67.69 & 65.73 & 63.13 & 65.97 & 43.86 & 66.55 & 68.24 & 54.89 & 63.40 & 63.07 & {6.63}\\
SkySenseGPT~\cite{skysensegpt} & ViT-L & 87.35 & 87.72 & 87.91 & 79.66 & 87.67 & 84.78 & 83.08 & 87.71 & 63.11 & 86.86 & 83.61 & 75.49 & 85.25 & 82.74 & {5.28}\\
VHM~\cite{vhm} & ViT-L & 80.60 & 79.81 & 80.67 & 76.10 & 80.82 & 78.81 & 76.92 & 79.55 & 59.74 & 80.47 & 75.43 & 72.57 & 79.73 & 76.72 & {\textbf{4.81}}\\
\bottomrule
\end{tabular}}
\label{tab_cls}
\end{table}
\vspace{-6pt}

\subsection{Semantic Segmentation}
The ISPRS Potsdam dataset provides both non-eroded and eroded labels, corresponding to annotations with and without object boundaries, respectively. In our experiments, we use the non-eroded labels for model evaluation and report the mean IoU for MIM- and CL-based methods. We omit LLM-based methods for semantic segmentation due to the lack of open-source MLLMs for this task. From Table \ref{tab_seg}, we can draw the following findings: 1) both MIM- and CL-based methods suffer from serious performance under image corruptions for the object detection task, with a mIoU drop of more than 10\%. 2) MIM-based methods achieve much better performance than CL-based methods on clean and noisy images. This is probably because MIM-based methods can capture local details by pixel reconstruction, while CL-based methods align global visual embeddings with text features, thus losing fine-grained details. Specifically, ScaleMAE~\cite{scalemae} achieves the best segmentation performance under image corruption, with an average mIoU of 60.02\%. 3) MIM-based methods suffer from a more serious performance drop under image corruptions than CL-based methods. Specifically, GeoRSCLIP~\cite{rs5m} achieves the best robustness across image corruptions, with a drop of 10.01\% of mIoU under corruptions. 

\vspace{-4pt}
\begin{table}[h]
\centering
\caption{Semantic segmentation performance (mIoU) on the ISPRS Potsdam dataset under different image perturbations.}
\small 
\setlength{\tabcolsep}{4pt} 
\resizebox{\textwidth}{!}{
\begin{tabular}{@{}lcccccccccccccccc@{}}
\toprule
\multirow{2}{*}{Method} & \multirow{2}{*}{Backbone} & \multirow{2}{*}{Clean} & Brightness & \multirow{2}{*}{Cloud} & Compression & Data & Gauss & Gauss & \multirow{2}{*}{Haze} & Motion & \multirow{2}{*}{Rotate} & Salt & \multirow{2}{*}{Scale} & \multirow{2}{*}{Translate} & \multirow{2}{*}{Avg} & \multirow{2}{*}{\(\mathcal{R}_{\text{TP}}\)}\\
 & & & Contrast & & Artifacts & Gaps & Blur & Noise &  & Blur &  & Pepper &  &  &  & \\
\hline
\multicolumn{16}{c}{MIM-based} \\
\hline
SatMAE~\cite{satmae} & ViT-L & 59.51 & 50.18 & 37.39 & 48.23 & 51.15 & 57.89 & 41.83 & 44.95 & 57.52 & 56.02 & 36.07 & 54.81 & 59.09 & 49.59 & {16.67}\\
ScaleMAE~\cite{scalemae} & ViT-L & 68.92 & 64.37 & 65.43 & \textbf{49.96} & 41.84 & 64.86 & \textbf{54.89} & \textbf{63.89} & 64.49 & 64.51 & \textbf{52.12} & 65.45 & 68.38 & \textbf{60.02} & {12.91}\\
RVSA~\cite{rvsa}  & ViT-B & \textbf{69.82} & \textbf{64.71} & \textbf{65.67} & 47.99 & 45.34 & \textbf{66.89} & 48.89 & 61.52 & \textbf{65.25} & \textbf{65.58} & 45.26 & \textbf{66.87} & \textbf{69.23} & 59.43 & {14.88}\\
SatMAE++~\cite{satmae++}  & ViT-L & 62.68 & 53.91 & 59.06 & 49.74 & \textbf{53.94} & 60.38 & 44.32 & 48.80 & 60.44 & 58.34 & 39.45 & 58.13 & 61.94 & 54.04 & {13.78}\\
\hline
\multicolumn{16}{c}{CL-based} \\
\hline
RemoteCLIP~\cite{remoteclip}  & ViT-B & 50.28 & 42.32 & 45.27 & 39.33 & 37.78 & 50.26 & 48.46 & 36.61 & 50.06 & 46.48 & 46.91 & 48.39 & 49.63 & 45.12 & {10.26}\\
RemoteCLIP~\cite{remoteclip}  & ViT-L & 56.69 & 54.51 & 52.73 & 43.24 & 51.19 & 50.82 & 45.12 & 50.47 & 51.82 & 53.68 & 38.98 & 54.59 & 56.53 & 50.31 & {11.25}\\
GeoRSCLIP~\cite{rs5m}  & ViT-B & 51.44 & 42.89 & 46.41 & 40.37 & 38.64 & 51.35 & 49.79 & 38.56 & 51.15 & 47.89 & 48.24 & 49.28 & 50.87 & 46.29 & {\textbf{10.01}}\\
GeoRSCLIP~\cite{rs5m}  & ViT-L & 56.81 & 54.97 & 52.53 & 43.28 & 41.37 & 50.49 & 42.7 & 49.41 & 51.36 & 53.54 & 36.98 & 54.66 & 56.64 & 48.99 & {13.77}\\
\bottomrule
\end{tabular}}
\label{tab_seg}
\end{table}
\vspace{-6pt}

\subsection{Object Detection}
We evaluate robustness on the corrupted images from the DIOR dataset. We report mAP for MIM- and CL-based methods. We omit LLM-based methods for the object detection task due to the lack of open-source LLMs for this task. From Table \ref{tab_det}, we can draw the following findings: 1) both MIM- and CL-based methods suffer from serious performance under image corruptions, with a mAP drop of more than 8\%. 2) MIM- and CL-based methods achieve comparable performance for object detection on clean and noisy images. 
RVSA~\cite{rvsa} attains the highest mAP of 70.96\% on clean images but experiences the most severe performance decline under image corruptions. 
3) MIM-based and CL-based methods exhibit a similar degree of performance degradation when subjected to image corruptions. 
Among them, SatMAE++~\cite{satmae++} demonstrates the most robust detection performance under noisy conditions.

\vspace{-4pt}
\begin{table}[h]
\centering
\caption{Object detection performance (mAP) on the DIOR dataset across different image perturbations.}
\small 
\setlength{\tabcolsep}{4pt} 
\resizebox{\textwidth}{!}{
\begin{tabular}{@{}lcccccccccccccccc@{}}
\toprule
\multirow{2}{*}{Method}  & \multirow{2}{*}{Backbone} & \multirow{2}{*}{Clean} & Brightness & \multirow{2}{*}{Cloud} & Compression & Data & Gauss & Gauss & \multirow{2}{*}{Haze} & Motion & \multirow{2}{*}{Rotate} & Salt & \multirow{2}{*}{Scale} & \multirow{2}{*}{Translate} & \multirow{2}{*}{Avg} & \multirow{2}{*}{\(\mathcal{R}_{\text{TP}}\)}\\
 & & & Contrast & & Artifacts & Gaps & Blur & Noise &  & Blur &  & Pepper &  &  &  & \\
\hline
\multicolumn{16}{c}{MIM-based} \\
\hline
SatMAE~\cite{satmae}  & ViT-L & 62.30 & 56.84 & 57.86 & 55.80 & 58.36 & 55.38 & 58.44 & 59.34 & 56.92 & 56.60 & 53.76 & 51.58 & 60.90 & 56.82 & {8.81}\\
ScaleMAE~\cite{scalemae}  & ViT-L & 70.20 & 64.80 & 65.98 & 62.50 & 64.46 & 62.58 & \textbf{63.82} & 66.10 & \textbf{63.08} & 63.44 & \textbf{60.50} & 53.08 & 68.26 & 63.22 & {9.94}\\
RVSA~\cite{rvsa}  & ViT-B & \textbf{70.96} & 60.59 & 65.02 & 61.58 & 64.60 & 62.35 & 62.87 & 63.98 & 62.88 & \textbf{64.04} & 56.61 & 55.97 & \textbf{69.69} & 62.51 & {11.91}\\
SatMAE++~\cite{satmae++}  & ViT-L & 65.20 & 59.44 & 61.02 & 60.30 & 59.88 & 59.66 & 61.06 & 61.72 & 59.56 & 59.14 & 58.64 & 48.48 & 64.70 & 59.47 & {\textbf{8.79}}\\
\hline
\multicolumn{16}{c}{CL-based} \\
\hline
RemoteCLIP~\cite{remoteclip}  & ViT-B & 60.40 & 56.72 & 56.28 & 56.78 & 54.56 & 53.68 & 57.36 & 55.90 & 53.42 & 54.54 & 54.40 & 44.92 & 59.72 & 54.86 & {9.17}\\
RemoteCLIP~\cite{remoteclip}  & ViT-L & 70.20 & \textbf{66.52} & \textbf{66.62} & 63.84 & \textbf{65.40} & \textbf{63.62} & 63.68 & \textbf{66.76} & 62.66 & 63.52 & 59.16 & \textbf{57.42} & 68.64 & 63.99 & {8.85}\\
GeoRSCLIP~\cite{rs5m}  & ViT-B & 60.20 & 56.28 & 56.04 & 56.08 & 55.46 & 53.38 & 56.92 & 55.50 & 53.38 & 53.98 & 53.48 & 46.98 & 59.32 & 54.73 & {9.09}\\
GeoRSCLIP~\cite{rs5m}  & ViT-L & 69.80 & 66.12 & 65.34 & \textbf{65.34} & 64.96 & \textbf{63.62} & 62.90 & 66.04 & 62.02 & 62.68 & 56.04 & 57.40 & 68.10 & \textbf{63.38} & {9.20}\\
\bottomrule
\end{tabular}}
\label{tab_det}
\end{table}
\vspace{-6pt}

\subsection{Image Captioning}
Table \ref{tab_cap} presents the zero-shot image captioning performance of GeoChat~\cite{geochat}, SkySenseGPT~\cite{skysensegpt}, VHM~\cite{vhm}, RS-LLaVA~\cite{rsllava}, and the recently introduced Falcon model~\cite{falcon}. Following the VRSBench protocol~\cite{vrsbench}, caption quality is evaluated using the GPT-4-based CLAIR metric~\cite{clair}\footnote{We use the \texttt{gpt-4o-mini-2024-07-18} model to compute the CLAIR scores.}. Given that geometric distortions—such as rotation, scaling, and translation—can substantially alter image content, we exclude performance measurements under these corruption conditions for image captioning, VQA, and visual grounding tasks. 

As shown in the upper part of Table~\ref{tab_cap}, all models experience performance degradation under noisy conditions. Among them, the Falcon~\cite{falcon} model achieves the best overall performance, significantly outperforming other methods on both clean and corrupted images. {However, it also suffers the largest performance drop of 6.28\%.}
In contrast, RS-LLaVA~\cite{rsllava} demonstrates the strongest robustness to image corruptions, exhibiting the smallest decrease in CLAIR score, with only a 2.03\% drop.
Additionally, we present results for GeoChat~\cite{geochat}, fine-tuned on the VRSBench training set, as shown in the lower part of Table \ref{tab_cap}. The fine-tuned GeoChat model on the target dataset exhibits significantly improved performance compared to its zero-shot counterpart, {with similar performance drop under corruptions.}

\vspace{-4pt}
\begin{table}[h]
\centering
\caption{Image captioning performance (CLAIR) on the VRSBench-Cap dataset across different image perturbations. \textit{ft} denotes models trained on the VRSBench training set.}
\small 
\setlength{\tabcolsep}{4pt} 
\resizebox{\textwidth}{!}{
\begin{tabular}{@{}lcccccccccccccc@{}}
\toprule
\multirow{2}{*}{Method} & \multirow{2}{*}{Backbone} & \multirow{2}{*}{Clean} & Brightness & \multirow{2}{*}{Cloud} & Compression & Data & Gauss & Gauss & \multirow{2}{*}{Haze} & Motion & Salt & \multirow{2}{*}{Avg} & \multirow{2}{*}{\(\mathcal{R}_{\text{TP}}\)}\\
 & & & Contrast & & Artifacts & Gaps & Blur & Noise &  & Blur & Pepper &  & \\
\midrule
GeoChat~\cite{geochat}  & ViT-L & 41.39 & 40.06 & 40.45 & 37.65 & 40.20 & 39.76 & 38.48 & 40.38 & 39.92 & 37.61 & 39.59 & {4.35}\\
SkySenseGPT~\cite{skysensegpt}  & ViT-L & 48.29 & 47.21 & 46.64 & 44.22 & 46.25 & 45.52 & 44.97 & 46.14 & 45.13 & 44.36 & 45.60 & {5.57}\\
VHM~\cite{vhm} & ViT-L & 52.02 & 50.19 & 50.82 & 50.26 & 50.57 & 51.22 & 50.46 & 50.39 & 50.72 & 49.48 & 50.46 & {3.00}\\
RS-LLaVA~\cite{rsllava} & ViT-L & 51.30 & 51.15 & 50.43 & 51.78 & 50.54 & 52.01 & 47.84 & 50.57 & 49.88 & 48.12 & 50.26 & {\textbf{2.03}}\\
Falcon~\cite{falcon}  & DaViT-B & \textbf{61.90} & \textbf{59.98} & \textbf{60.09} & \textbf{57.13} & \textbf{59.48} & \textbf{57.43} & \textbf{56.31} & \textbf{59.85} & \textbf{59.94} & \textbf{51.83} & \textbf{58.01} & {6.28}\\
\midrule
\rowcolor{gray!20}
GeoChat$_{ft}$~\cite{geochat}  & ViT-L & 71.26 & 69.00 & 68.93 & 66.60 & 69.45 & 68.63 & 67.83 & 69.98 & 69.02 & 63.87 & 68.15 & {4.36}\\
\bottomrule
\end{tabular}}
\label{tab_cap}
\end{table}
\vspace{-6pt}

\subsection{Visual Question Anaswering}

Table \ref{tab_vqa} reports VQA performance across various image perturbations. Following the VRSBench protocol~\cite{vrsbench}, VQA performance is evaluated using the GPT-4-based matching accuracy\footnote{We use the \texttt{gpt-4o-mini-2024-07-18} model to compute the matching accuracy for VQA.}. From Table~\ref{tab_vqa}, it is evident that all LLM-based models experience a moderate decline in performance under image perturbations. Overall, VHM~\cite{vhm} achieves the best accuracy across both clean and noisy images. LHRS-Bot~\cite{lhrs}, RS-LLaVA~\cite{rsllava}, and Falcon~\cite{falcon}, despite showing relatively lower overall accuracy, exhibit less sensitivity to image corruptions. Additionally, the GeoChat~\cite{geochat} fine-tuned on the VRSBench training set surpasses zero-shot models in terms of absolute performance, {while exhibiting a slightly smaller performance drop under perturbations, indicating improved robustness.}

\vspace{-4pt}
\begin{table}[ht]
\centering
\caption{VQA performance (Accuracy) on the VRSBench-VQA dataset across different image perturbations. \textit{ft} indicates models fine-tuned on the VRSBench training set.}
\small 
\setlength{\tabcolsep}{4pt} 
\resizebox{\textwidth}{!}{
\begin{tabular}{@{}lcccccccccccccc@{}}
\toprule
\multirow{2}{*}{Method} & \multirow{2}{*}{Backbone} & \multirow{2}{*}{Clean} & Brightness & \multirow{2}{*}{Cloud} & Compression & Data & Gauss & Gauss & \multirow{2}{*}{Haze} & Motion & Salt & \multirow{2}{*}{Avg} & \multirow{2}{*}{\(\mathcal{R}_{\text{TP}}\)}\\
 & & & Contrast & & Artifacts & Gaps & Blur & Noise &  & Blur & Pepper &  & \\
\midrule
GeoChat~\cite{geochat}  & ViT-L & 56.63 & 53.89 & 54.82 & 55.14 & 55.99 & 55.88 & 55.44 & 56.22 & 54.08 & 54.04 & 55.06 & {2.77}\\
LHRS-Bot~\cite{lhrs}  & ViT-L & 35.72 & 35.72 & 35.69 & 35.72 & 35.72 & 35.72 & 35.72 & 35.72 & 35.34 & 35.72 & 35.56 & {\textbf{0.45}}\\
SkySenseGPT~\cite{skysensegpt}  & ViT-L & 60.21 & 59.26 & 59.73 & 57.93 & 59.64 & 59.21 & 58.27 & 59.63 & 59.17 & 57.27 & 58.90 & {2.18}\\
VHM~\cite{vhm} & ViT-L & \textbf{61.72} &  \textbf{60.91} &  \textbf{61.07} &  \textbf{60.40} &  \textbf{61.49} &  \textbf{60.91} &  \textbf{60.91} &  \textbf{61.12} &  \textbf{59.97} &  \textbf{60.39} &  \textbf{60.90} & {1.33}\\
RS-LLaVA~\cite{rsllava} & ViT-L & 57.25 & 57.04 & 57.14 & 55.45 & 57.25 & 57.14 & 55.97 & 57.21 & 55.25 & 55.82 & 56.47 & {1.36}\\
Falcon~\cite{falcon}  &  DaViT-B & 33.27 & 32.83 & 32.70 & 32.19 & 33.30 & 33.43 & 32.85 & 32.76 & 32.97 & 31.55 & 32.73 & {1.59}\\
\midrule
\rowcolor{gray!20}
GeoChat$_{ft}$~\cite{geochat}  & ViT-L & 75.79 & 75.13 & 74.97 & 73.84 & 75.63 & 74.89 & 74.46 & 75.43 & 74.76 & 72.77 & 74.65 & {1.50}\\
\bottomrule
\end{tabular}}
\label{tab_vqa}
\end{table}
\vspace{-6pt}

\subsection{Visual Grounding}

Table \ref{tab_vg} presents the zero-shot visual grounding performance of comparing methods. We report grounding accuracy at an IoU threshold of 0.5. As shown in the upper part of Table~\ref{tab_vg}, all methods experience noticeable declines in performance under image perturbations. The GeoGround~\cite{geoground} model achieves the best performance on both clean and perturbed images, with a grounding accuracy of 75.93\% {and the smallest drop of 4.48\%.} In comparison, the Falcon~\cite{falcon} model, despite not being trained on VRSBench, demonstrates competitive visual grounding capability, but experiences a more significant degradation in performance when exposed to image corruptions. The fine-tuned GeoChat~\cite{geochat} model shows substantial improvements over its zero-shot counterpart{, with a substantially reduced performance drop under noisy conditions.}

\vspace{-4pt}
\begin{table}[h]
\centering
\caption{Visual grounding performance on the VRSBench-Ref dataset across different image perturbations. We report grounding accuracy at an IoU threshold of 0.5. * indicates the GeoGround model includes VRSBench in its training data. \textit{ft} indicates models fine-tuned on the VRSBench training set.}
\small 
\setlength{\tabcolsep}{4pt} 
\resizebox{\textwidth}{!}{
\begin{tabular}{@{}lcccccccccccccc@{}}
\toprule
\multirow{2}{*}{Method} & \multirow{2}{*}{Backbone} & \multirow{2}{*}{Clean} & Brightness & \multirow{2}{*}{Cloud} & Compression & Data & Gauss & Gauss & \multirow{2}{*}{Haze} & Motion & Salt & \multirow{2}{*}{Avg} & \multirow{2}{*}{\(\mathcal{R}_{\text{TP}}\)}\\
 & & & Contrast & & Artifacts & Gaps & Blur & Noise &  & Blur & Pepper &  & \\
\midrule
GeoChat~\cite{geochat}  & ViT-L & 18.96 & 17.09 & 16.54 & 16.52 & 16.19 & 16.61 & 16.93 & 17.09 & 16.91 & 16.57 & 16.72 & {11.81}\\
VHM~\cite{vhm} & ViT-L & 37.20 & 34.66 & 35.29 & 34.18 & 35.48 & 35.01 & 35.78 & 35.54 & 32.21 & 34.58 & 34.74 & {6.61}\\
GeoGround$^{*}$~\cite{geoground}  & ViT-L & \textbf{75.93} & \textbf{73.57} & \textbf{71.57} & \textbf{71.30} & \textbf{72.23} & \textbf{73.23} & \textbf{72.92} & \textbf{74.06} & \textbf{72.11} & \textbf{71.77} & \textbf{72.53} & {\textbf{4.48}}\\
Falcon~\cite{falcon} &  DaViT-B & 73.30 & 71.31 & 69.92 & 65.83 & 68.61 & 70.79 & 64.28 & 71.04 & 68.17 & 59.53 & 67.72 & {7.61}\\
\midrule
\rowcolor{gray!20}
GeoChat$_{ft}$~\cite{geochat}  & ViT-L & 55.50 & 53.79 & 52.20 & 50.51 & 53.06 & 53.11 & 51.57 & 54.23 & 52.99 & 49.82 & 52.36 & {5.66}\\
\bottomrule
\end{tabular}}
\label{tab_vg}
\end{table}
\vspace{-6pt}

\vspace{-4pt}
\section{Discussion}
In this section, we further analyze the robustness of EOFMs across model architectures, tasks, corruption categories, and backbone sizes.

\subsection{Vision-Centric vs. Vision-Language Foundation Models}
As shown in Fig.~\ref{fig_diss_task}, vision-centric foundation models (MIM-based) tend to suffer greater performance degradation under visual perturbations compared to vision-language models (CL- and VLM-based). This difference is especially pronounced in image-level scene classification tasks, where MIM-based models exhibit an average performance drop exceeding 25\%. In contrast, vision-language models consistently demonstrate stronger robustness across tasks, maintaining performance drops below 10\% in most cases. This is probably due to the complementary grounding effect of language supervision. We also note that the robustness gap between vision-centric and vision-language models is less significant for segmentation and detection tasks.

\subsection{Robustness Across Different Tasks}
\begin{figure}[ht!]
    \centering
    \begin{minipage}[c]{0.4\linewidth}
        \begin{minipage}[t]{\linewidth}
            \centering
            \includegraphics[width=\linewidth,height=3.5cm]{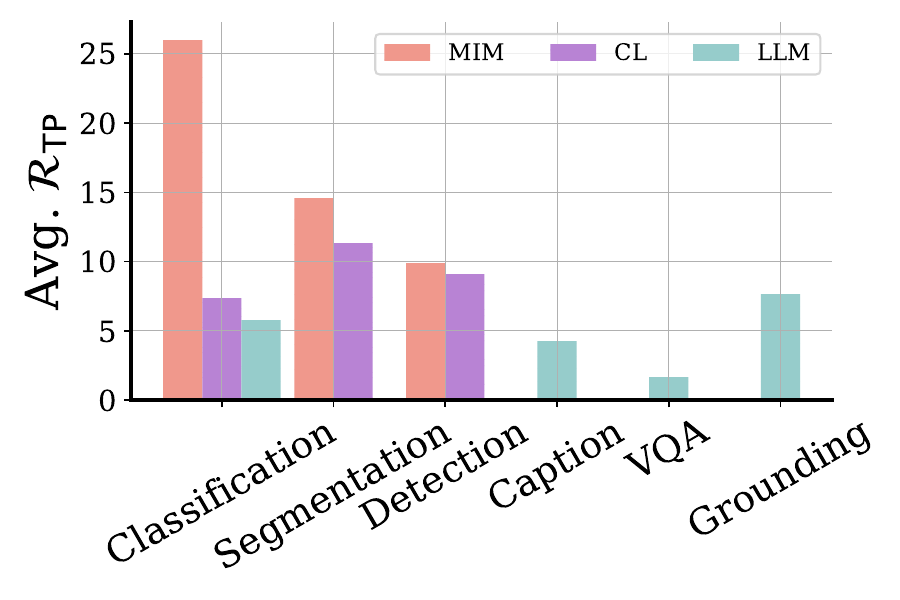}
            \caption{{Robustness across different tasks and model architectures. We report the average \(\mathcal{R}_{\text{TP}}\) across models.}}
            \label{fig_diss_task}
        \end{minipage}
        \vspace{0.2cm}

        \begin{minipage}[b]{\linewidth}
            \centering
            \includegraphics[width=\linewidth,height=3.5cm]{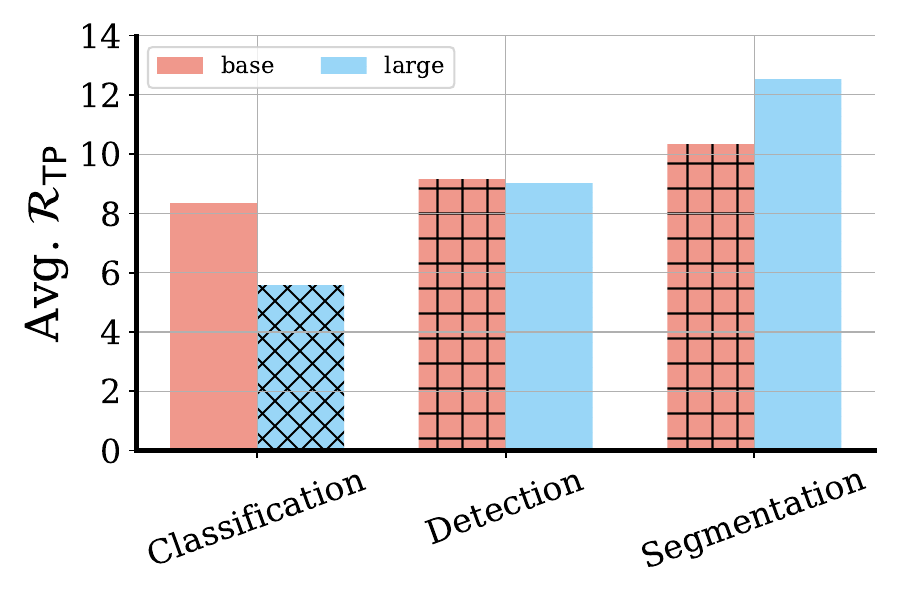}
            \caption{Robustness across different backbone sizes. We report the average \(\mathcal{R}_{\text{TP}}\) for RemoteCLIP and GeoRSCLIP.}
            \label{fig_diss_backbone}
        \end{minipage}
    \end{minipage}%
    \hfill
    \begin{minipage}[c]{0.55\linewidth}
        \centering
        \includegraphics[width=\linewidth,height=8.3cm]{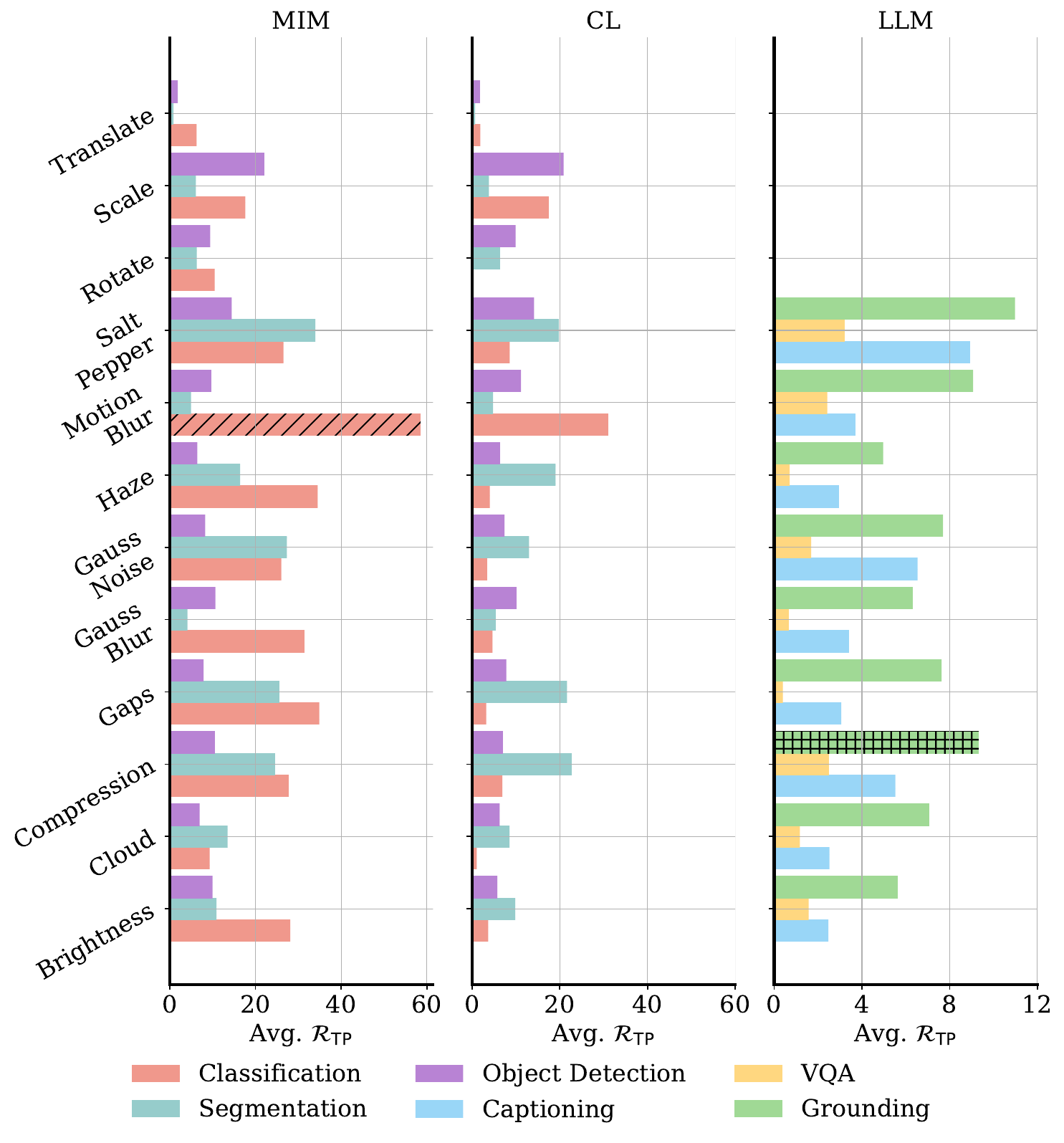}
        \caption{Robustness across different types of corruptions. We report the  \(\mathcal{R}_{\text{TP}}\) across models.}
        \label{diss_noise}
    \end{minipage}
\end{figure}

Fig.~\ref{fig_diss_task} further highlights that vulnerability to perturbations varies substantially across tasks. MIM-based models are particularly sensitive in classification tasks, while CL-based models maintain greater stability across classification, segmentation, and detection tasks. This can be attributed to the contrastive objective, which encourages learning of invariant and robust representations. LLM-based models, on the other hand, show the smallest performance degradation in vision-language tasks such as image captioning and visual question answering (VQA)—typically below 5\%. These results suggest that LLM-based methods excel in corruption-robust generalization, particularly in tasks that benefit from multimodal alignment.

\subsection{Robustness Across Different Backbone Sizes}
For scene classification, the larger backbone (ViT-L) demonstrates greater robustness, showing less performance degradation under image corruptions. However, for fine-grained tasks such as semantic segmentation and object detection, ViT-L suffers larger performance drops compared to ViT-B. This suggests that while a larger backbone may enhance robustness in high-level recognition tasks, it may also amplify sensitivity to image corruptions in pixel- or region-level tasks.

\subsection{Robustness Across Perturbation Types}
As shown in Fig.~\ref{diss_noise}, the performance degradation of MIM-, CL-, and LLM-based models varies notably across different visual perturbations. Motion blur causes the most severe drop, especially for MIM, which loses around {60}\% in performance, indicating a high sensitivity to spatial distortions. In contrast, translation has the least impact, suggesting minimal disruption to pattern recognition. Across perturbation types, LLM-based models consistently exhibit the strongest robustness, maintaining performance drops below {10}\% in most cases. This reinforces the value of language supervision in promoting the learning of more semantic and perturbation-invariant features.

\subsection{Robustness Across Compound Perturbations}
Real-world images often suffer from multiple simultaneous degradations (e.g., haze combined with noise). To investigate this, we evaluate object detection performance under selected compound perturbations on the DIOR-R dataset. The results from Table~\ref{tab_compound} reveal that models exhibit substantially lower accuracy under compound perturbations compared to single perturbations. For example, the detection performance of RemoteCLIP drops from 56.72\% under Brightness to 40.58\% under all three combined perturbations. Moreover, certain combinations of perturbations, such as \emph{Brightness + Compression}, produce synergistic effects, causing performance declines that exceed the sum of individual perturbation effects.

\begin{table}[h!]
\centering
\caption{Object Detection performance (mAP) on DIOR-R under compound corruptions}
\small 
\setlength{\tabcolsep}{4pt} 
\resizebox{\textwidth}{!}{%
\begin{tabular}{lcccccccc}
\toprule
\multirow{2}{*}{Model} & \multirow{2}{*}{Clean} & \multirow{2}{*}{Brightness} & \multirow{2}{*}{Clouds} & \multirow{2}{*}{Compression} & Brightness & Brightness& Clouds & \multirow{2}{*}{All Three} \\
 & & & & & + Clouds & + Compression & + Compression & \\
\midrule
Brightness  & \icono     & \icoyes  & \icono     & \icono     & \icoyes  & \icoyes  & \icono     & \icoyes \\
Clouds      & \icono     & \icono    & \icoyes   & \icono     & \icoyes  & \icono    & \icoyes   & \icoyes \\
Compression & \icono     & \icono    & \icono     & \icoyes   & \icono    & \icoyes  & \icoyes   & \icoyes \\
\midrule
RemoteCLIP  & 60.40 & 56.72 & 56.28 & 56.78 & 49.98 & 45.36 & 52.57 & 40.58 \\
GeoRSCLIP   & 60.20 & 56.28 & 56.04 & 56.08 & 50.27 & 44.90 & 52.39 & 40.94 \\
\bottomrule
\end{tabular}%
} 
\label{tab_compound}
\vspace{-6pt}
\end{table}

\subsection{Robustness in Multispectral Remote Sensing}
In addition to high-resolution RGB imagery, multispectral data provide complementary spectral information that can improve scene understanding and robustness in Earth Observation. 
To assess the robustness of EO foundation models beyond optical imagery, we conduct preliminary experiments on two representative multispectral datasets: fMoW-Sentinel2~\cite{christie2018functional} and BigEarthNet~\cite{sumbul2019bigearthnet}, using two recently proposed foundation models, SatMAE~\cite{satmae} and EarthDial~\cite{soni2025earthdial}. 
The results, summarized in Table~\ref{tab_multispectral}, show substantial performance drops under image perturbations, 
underscoring the brittleness of current EO foundation models when applied to multispectral data.

\vspace{-4pt}
\begin{table}[h]
\centering
\caption{Scene classification performance on the multispectral dataset across different image perturbations.}
\small 
\setlength{\tabcolsep}{4pt} 
\resizebox{\textwidth}{!}{
\begin{tabular}{@{}lcccccccccccccccc@{}}
\toprule
\multirow{2}{*}{Method}  & \multirow{2}{*}{Backbone} & \multirow{2}{*}{Clean} & Brightness & \multirow{2}{*}{Cloud} & Compression & Data & Gauss & Gauss & \multirow{2}{*}{Haze} & Motion & \multirow{2}{*}{Rotate} & Salt & \multirow{2}{*}{Scale} & \multirow{2}{*}{Translate} & \multirow{2}{*}{Avg} & \multirow{2}{*}{\(\mathcal{R}_{\text{TP}}\)}\\
 & & & Contrast & & Artifacts & Gaps & Blur & Noise &  & Blur &  & Pepper &  &  &  & \\
\midrule
SatMAE~\cite{satmae}  & fMoW-S2~\cite{christie2018functional} & 59.75 & 37.46 & 58.69 & 33.58 & 50.01 & 29.35 & 36.64 & 41.57 & 40.12 & 38.93 & 51.79 & 43.35 & 59.68 & 43.43 & 27.31 \\
EarthDial~\cite{soni2025earthdial} & BigEarthNet~\cite{sumbul2019bigearthnet} & 46.521 & 31.47 & 46.48 & 45.40 & 34.97 & 36.37 & 38.69 & 33.25 & 39.84 & 29.51 & 22.71 & 39.18 & 45.30 & 36.93 & 20.62 \\
\bottomrule
\end{tabular}}
\label{tab_multispectral}
\end{table}

\section{Related Works}

\textbf{Robustness Research in Remote Sensing.} Deep learning (DL)-based methods have achieved significant success in remote sensing image processing; however, their black-box nature raises concerns regarding interpretability, transparency, and vulnerability to adversarial examples. Recent studies have begun to address the robustness of DL models in this domain. Kazmi et al.\cite{Kazmi2023AdversarialAO} present a comprehensive literature review on adversarial attacks in aerial imagery processing, but do not provide an in-depth analysis of model robustness. Mei et al.\cite{Mei2023ACS} examine the robustness of DL-based methods for remote sensing image understanding, with a focus on image classification and object detection tasks. Lian et al.~\cite{lian2023cba,lian2022benchmarking} propose techniques to enhance adversarial robustness specifically for object detection in aerial imagery. \cite{Tasneem2024ImproveAR} aims to improve the adversarial robustness of scene classification models in remote sensing via CAM-guided feature learning. These works only study the robustness of task-specific models. In contrast, our work for the first time investigates the robustness of foundation models in remote sensing.

\textbf{Foundation Models in Remote Sensing.} In general, there are four types of foundation models in remote sensing: MIM-based, CL-based, LLM-based, and diffusion-based methods. (1) Masked Image Modeling (MIM) has gained popularity through the pioneering work of MAE~\cite{mae}. These methods typically employ an encoder network to learn feature representations by masking a portion of the image tokens, followed by a decoder network that reconstructs the masked image pixels in a self-supervised manner. In the field of remote sensing, notable approaches include SatMAE~\cite{satmae}, RingMo~\cite{ringmo}, RVSA~\cite{rvsa}, ScaleMAE~\cite{scalemae}, SatMAE++~\cite{satmae++}, and DOFA~\cite{dofa}. (2) Contrastive Learning (CL) employs separate encoders to project images and texts into a shared embedding space, using a contrastive objective to align the resulting embeddings. Building on the success of the pioneering work CLIP~\cite{clip}, several contrastive learning-based foundation models have been introduced in the field of remote sensing, including RS-CLIP~\cite{rs5m}, RemoteCLIP~\cite{remoteclip}, GeoRSCLIP~\cite{rs5m}, SkyCLIP~\cite{skyclip}, S-CLIP~\cite{s-clip}, SatCLIP~\cite{satclip}, and GeoCLIP~\cite{geoclip}. (3) Following the pioneering works of MiniGPT-4~\cite{minigpt4} and LLaVA~\cite{llava}, multimodal large language models (MLLMs) have attracted significant research attention in recent years. For instance, RSGPT~\cite{rsgpt} introduces the first GPT-based MLLM tailored for remote sensing image understanding. Other notable approaches include GeoChat~\cite{geochat}, EarthGPT~\cite{earthgpt}, EarthMarker~\cite{earthmarker}, Popeye~\cite{popeye}, RS-LLaVA~\cite{rsllava}, VHM~\cite{vhm}, LHRS-Bot~\cite{lhrs}, SkyEyeGPT~\cite{skyeyegpt}, SkySenseGPT~\cite{skysensegpt}, RSUniVLM~\cite{liu2024rsunivlm}, and Falcon~\cite{falcon}. (4) Diffusion-based foundation models learn the joint distribution between text prompts and images through a forward noising process followed by a reverse denoising process. Recent studies have applied these models to synthesize satellite~\cite{Khanna2023DiffusionSatAG,Liu2025Text2EarthUT}, aerial~\cite{Arrabi2024CrossViewMD}, hyperspectral~\cite{Pang2024HSIGeneAF}, and multi-resolution imagery~\cite{Yu2024MetaEarthAG}. 

\section{Conclusion and Future Work}
In this work, we present \papernameAbbrev, the first comprehensive benchmark for evaluating the robustness of EOFMs across six core tasks and twelve perturbation types. Our evaluation reveals that existing EOFMs experience noticeable performance degradation under image corruptions. We also observe significant variations in robustness across model types, task categories, and backbone sizes, offering valuable insights for future development of robust models. We hope \papernameAbbrev{} will serve as a standard benchmark to drive the creation of more robust and reliable models for Earth observation.


Despite its contributions, this work has several limitations. First, the evaluation is limited to high-resolution optical imagery, excluding other key modalities such as multispectral (e.g., Sentinel-2), hyperspectral, and SAR data. Second, the benchmark’s dataset and task coverage are not exhaustive. While it includes widely used datasets (AID, Potsdam, DIOR, VRSBench), they may not fully reflect global variation in geography, resolution, or sensor types. Additionally, important tasks such as change detection, region captioning, and object counting are currently not included.

\section{Broader Impact}
\papernameAbbrev{} aims to improve the reliability of Earth observation foundation models by systematically evaluating their robustness to real-world noise and perturbations. This is critical for high-stakes applications such as disaster response and environmental monitoring. By identifying vulnerability patterns across tasks and models, our benchmark can guide the development of future robust models.

\begin{acksection}
This work is supported by the EDF-KOIOS project and used the Dutch national e-infrastructure with the support of the SURF Cooperative using the funding of the projects EINF-12538, EINF-10925 and NWO-2023.027. We would like to express our deepest gratitude to the anonymous reviewers whose insightful comments and suggestions significantly improved the quality of this paper. 
\end{acksection}

\newpage
{
\small
\bibliographystyle{unsrt}
\bibliography{egbib} 
}

\newpage
\newpage
\appendix




\section{{\papernameAbbrev} Documentation and Intended Uses}

\subsection{Overview}
{\papernameAbbrev} dataset includes four subsets: AID for scene classification, ISPRS Potsdam for semantic segmentation, DIOR for object detection, and VRSBench for image captioning, visual question answering (VQA), and visual grounding.

\subsection{Data Organization}
Our {\papernameAbbrev} dataset is organized as follows.

\dirtree{%
.1 root/.
.2 AID.
.3 AID\_train.zip.
.3 AID\_test.zip.
.3 AID\_JSON.
.2 Potsdam.
.3 Potsdam\_Images\_trian.zip.
.3 Potsdam\_Anns\_trian.zip.
.3 Potsdam\_Images\_test.zip.
.3 Potsdam\_Anns\_test.zip.
.2 DIOR.
.3 DIOR\_Images\_trian.zip.
.3 DIOR\_Anns\_trian.zip.
.3 DIOR\_Images\_test.
.4 clean.zip; brightness\_contrast.zip; cloud.zip; compression\_artifacts.zip....
.3 DIOR\_Anns\_test.
.4 clean.zip; rotate.zip; scale.zip; translate.zip.
.2 VRSBench.
.3 VRSBench\_Images\_trian.zip.
.3 VRSBench\_train.json.
.3 VRSBench\_Images\_test.
.4 clean.zip; brightness\_contrast.zip; cloud.zip; compression\_artifacts.zip....
.3 VRSBench\_EVAL\_Cap.json.
.3 VRSBench\_EVAL\_referring.json.
.3 VRSBench\_EVAL\_vqa.json.
}

Detailed descriptions for each folder or file are given below.
\begin{itemize}
    \item AID/Images\_train.zip contains all AID images in the training set.
    \item AID/Images\_test.zip contains images in the test set under corruption.
    \item AID/AID\_JSON folder contains json file for zero-shot evaluation of LLM-based models.
    \item Potsdam/Potsdam\_Images\_trian.zip contains all Potsdam images in the training set.
    \item Potsdam/Potsdam\_Images\_trian.zip contains all Potsdam images in the test set under corruption.
    \item Potsdam/Potsdam\_Anns\_trian.zip contains annotations for images in the training set.
    \item Potsdam/Potsdam\_Anns\_test.zip contains annotations for images in the test set under corruption.
    \item DIOR/DIOR\_Images\_trian.zip contains all DIOR images in the training set.
    \item DIOR/DIOR\_Anns\_trian.zip contains all Oriented Bounding Boxes annotations for images in the training set.
    \item DIOR/DIOR\_Images\_test folder contains all DIOR images in the test set under corruption.
    \item DIOR/DIOR\_Anns\_test folder contains oriented bounding box annotations for test images under four settings: clean, and three spatial transformations — rotate, scale, and translate. For corruptions that do not involve spatial transformations (e.g., blur, noise), annotations from the clean setting are reused, as these corruptions do not alter object positions or shapes.
    \item VRSBench/Annotation\_Images\_train.zip contains VRSBench images in the training set.
    \item VRSBench/Annotation\_Images\_test folder contains VRSBench images in the test set, one folder per noise type.
    \item VRSBench/VRSBench\_train.json contains VRSBench training annotations following LLaVA in standard JSON format.
    \item VRSBench/VRSBench\_EVAL\_Cap.json contains VRSBench evaluation annotations for the captioning task in standard JSON format.
    \item VRSBench/VRSBench\_EVAL\_referring.json contains VRSBench evaluation annotations for the visual grounding task in standard JSON format.
    \item VRSBench/VRSBench\_EVAL\_vqa.json contains VRSBench evaluation annotations for the VQA task in standard JSON format. 
\end{itemize}

\subsection{Intended Uses}

{\papernameAbbrev} is intended for use in academic and research settings, specifically for:
\begin{itemize}
    \item Evaluating the robustness of remote sensing foundation models.
    \item Understand the robustness of remote sensing foundation models across tasks, noise types, and model architectures.
\end{itemize}

\subsection{Use Cases}

\begin{itemize}
    \item \textbf{Robustness Research}: \papernameAbbrev{} provides a standardized testbed for evaluating the robustness of remote sensing foundation models under a wide range of realistic corruptions, making it valuable for both academic and applied robustness research.
    
    \item \textbf{Model Diagnosis and Comparison}: The benchmark enables fine-grained analysis of performance degradation across different model types (e.g., MIM-based, CL-based, LLM-based), tasks, and corruption types, serving as a practical tool for diagnosing model vulnerabilities and comparing architectures.
    
    \item \textbf{Guidance for Model Development}: Insights derived from \papernameAbbrev{} can inform the design of more resilient remote sensing models and training strategies, particularly for mission-critical applications such as disaster response and environmental monitoring.
    
\end{itemize}

\subsection{Limitations}

\begin{itemize}
    \item \textbf{Modal and Sensor Scope}: This benchmark focuses exclusively on optical remote sensing imagery and does not currently include multispectral, hyperspectral, or SAR modalities, limiting its applicability to other sensing systems.
    
    \item \textbf{Task Coverage}: While \papernameAbbrev{} includes six core tasks, it does not encompass important tasks such as change detection, region captioning, and object counting, which are increasingly relevant in Earth observation.

    \item \textbf{Dataset Representativeness}: The benchmark is constructed from well-known datasets such as AID, DIOR, Potsdam, and VRSBench, which may not fully capture the geographic, temporal, and sensor diversity of real-world global remote sensing data.

\end{itemize}

\subsection{Ethical Considerations}

\begin{itemize}
    \item \textbf{Public Data Usage}: \papernameAbbrev{} is built entirely from publicly available datasets. No personal, private, or sensitive information is included, and care has been taken to respect privacy in all sourced imagery.
    \item \textbf{Responsible Use}: We encourage the research community to use this benchmark ethically, particularly in downstream applications involving environmental monitoring, urban analytics, and decision-making systems that may influence public policy or resource allocation.

\end{itemize}

\subsection{Documentation and Maintenance}

\begin{itemize}
    \item \textbf{Versioning}: Detailed version history of the dataset will be maintained to track changes and improvements over time.

    \item \textbf{Version Control}: The benchmark will be versioned to ensure transparency in updates, including bug fixes, additional corruption types, and task expansions.
    
    \item \textbf{Community Feedback}: We welcome contributions and suggestions from the community to expand, refine, and validate the benchmark, fostering collaborative progress in building robust AI systems for Earth observation.

\end{itemize}

\subsection{Accountability Framework}

To promote responsible AI development, \papernameAbbrev{} adopts an open accountability model. Users are encouraged to report any issues related to dataset quality, annotation errors, or robustness evaluation inconsistencies. These reports will be reviewed and integrated into future updates, supporting a continuous feedback loop for improving the benchmark’s accuracy, fairness, and utility.

\section{URL to Data and Metadata}
The {\papernameAbbrev} dataset can be accessed and downloaded through our Huggingface repository (\url{\huggingpage}). Detailed metadata for the dataset is documented using the Croissant metadata framework, ensuring comprehensive coverage and compliance with the MLCommons Croissant standards, check [metadata](\url{https://huggingface.co/api/datasets/xiang709/REOBench}).

\section{Author Statement and Data License}

\textbf{Author Responsibility Statement:} The authors bear all responsibilities in case of any violations of rights or ethical concerns regarding the {\papernameAbbrev} dataset. 

\textbf{Data License Confirmation:} The dataset is released under the [CC-BY-4.0], which permits unrestricted use, distribution, and reproduction in any medium, provided the original work is properly cited.

\section{Hosting and Accessibility} The {\papernameAbbrev} dataset is hosted on Huggingface (\url{\huggingpage}) to ensure reliable and continuous accessibility. 

\textbf{Maintenance Plan:} Ongoing maintenance and updates will be managed by the dataset authors, with updates scheduled bi-annually or as significant changes in the data sources occur.

\textbf{Long-term Preservation:} The dataset is archived in Huggingface (\url{\huggingpage}) to ensure long-term availability.

\textbf{Structured Metadata:} The annotation for each image is well-organized in standard JSON format to ensure easy usage.

\section{Dataset Collection Details}

\subsection{Source datasets}
Our \papernameAbbrev{} uses three source datasets, i.e., AID for scene classification, ISPRS Potsdam for semantic segmentation, DIOR for object detection, and VRSBench for image captioning, visual question answering (VQA), and visual grounding. The details of each dataset are shown in Table~\ref{tab_source}.

\begin{table}[!ht]
    \centering
    \caption{Statistics of source datasets.}
    \begin{tabular}{l|cccc}
        \toprule
        Dataset & Train & Test & Category & Size \\ 
        \hline
        AID & 8,000 & 2,000 & 30 & $256\times256$ \\ 
        Potsdam & 3,456 & 2,016 & 6 & $512\times512$ \\  
        DIOR-R & 11,725 & 11,738 & 20 & $800\times800$ \\
        VRSBench & 20,264 & 9,350 & 26 & $512\times512$ \\ 
        \bottomrule
    \end{tabular}
    \label{tab_source}
\end{table}

\subsection{Image Pertubations}

\begin{table}[ht]
\centering
\caption{Configurations of different image corruption types and corresponding severity.}
\resizebox{0.98\textwidth}{!}{
\begin{tabular}{llccccc}
\hline
\textbf{Corruption Type} & \textbf{Parameter} & \textbf{S1} & \textbf{S2} & \textbf{S3} & \textbf{S4} & \textbf{S5} \\ \hline
Gaussian Noise & $\sigma$ & 0.04 & 0.05 & 0.06 & 0.07 & 0.08 \\ 
Salt Pepper Noise & amount & 0.005 & 0.01 & 0.02 & 0.03 & 0.05 \\
Gaussian Blur & kernel size & 3×3 & 5×5 & 7×7 & 9×9 & 11×11 \\
Motion Blur & kernel size & 2×2 & 4×4 & 6×6 & 8×8 & 10×10 \\
Brightness/Contrast & b / c & +0.0 / 1.0 & +0.1 / 0.8 & +0.2 / 0.6 & +0.3 / 0.4 & +0.4 / 0.2 \\
Clouds & threshold & 0.90 & 0.85 & 0.80 & 0.75 & 0.70 \\
Haze & intensity & 0.20 & 0.30 & 0.40 & 0.50 & 0.60 \\
Data Gaps & num / width (px) & 2 / 3 & 3 / 4 & 4 / 5 & 5 / 6 & 6 / 7 \\
Compression Artifacts & JPEG quality & 30 & 25 & 20 & 15 & 10 \\
Rotation & angle ($^\circ$) & 30 & 45 & 60 & 75 & 90 \\
Scaling & scale ratio & 0.9 & 0.8 & 0.7 & 0.6 & 0.5 \\
Translation & displacement (px) & ±15 & ±20 & ±25 & ±30 & ±35 \\
\hline
\end{tabular}
}
\label{tab_perturbations}
\end{table}

In table \ref{tab_perturbations}, we report parameter configures for different noise levels and types. S1 to S5 represent the severity levels of each corruption, with S1 being the mildest and S5 the most severe. The ``Parameter'' column lists the key control variables for each type of image corruption. For Gaussian noise, $\sigma$ represents the standard deviation that controls the amplitude of the Gaussian noise; for salt-and-pepper noise, the \texttt{amount} specifies the proportion of pixels randomly replaced with black or white; Gaussian blur and motion blur use \texttt{kernel size} to define the size of the convolution kernel; brightness and contrast corruption is controlled using $b / c$ to represent brightness offset and contrast scaling respectively; for the cloud corruption, \texttt{density} denotes the cloud coverage, which is controlled based on the Perlin noise threshold~\cite{touti2017perlin}; the parameter for haze simulation is \texttt{intensity}, a blending ratio that determines the degree of mixing with a white layer, and in subsequent visual fidelity experiments we extend the severity levels up to 9 with corresponding intensity values of 0.70, 0.80, 0.90, and 0.95 for S6–S9, respectively; data gaps are defined by \texttt{number/width of stripes}, which specifies the number and width of missing regions; Compression Artifacts are governed by the \texttt{JPEG quality} parameter, with lower values indicating stronger compression artifacts. In terms of geometric transformations, rotation is defined by \texttt{angle}, scaling is represented by \texttt{scale ratio}, and translation is described by the maximum offset in \texttt{displacement}. These parameters allow for systematic control and severity grading of each corruption type.

\subsection{Perturbation Fidelity Analysis}
To assess the visual fidelity of our synthetic perturbations, we compute the Fréchet Inception Distance (FID)~\cite{heusel2017gans} between corrupted images and their corresponding clean counterparts. Specifically, we report FID scores for four benchmarks with different image resolutions: DIOR (800$\times$800), VRSBench (512$\times$512), Potsdam (512$\times$512), and AID (256$\times$256). For each dataset, FID was calculated across five severity levels (1–5) of perturbations, using the clean dataset as the reference distribution. 
Table~\ref{tab_fid} shows that FID values remain consistently low across severity levels, suggesting that the corrupted datasets are visually similar to real-world imagery and therefore suitable for robustness evaluation. 

\begin{table}[ht]
\centering
\caption{FID scores between corrupted datasets and their corresponding clean datasets across different severity levels. Lower values indicate higher visual similarity.}
\begin{tabular}{c c c c c}
\toprule
Severity & DIOR (800$\times$800) & VRSBench (512$\times$512) & Potsdam (512$\times$512) & AID (256$\times$256) \\
\midrule
1 & 2.36  & 3.84  & 6.06  & 14.07 \\
2 & 5.46  & 7.67  & 10.35 & 23.60 \\
3 & 9.67  & 11.69 & 17.06 & 32.37 \\
4 & 14.21 & 14.76 & 24.86 & 39.31 \\
5 & 19.81 & 20.06 & 37.89 & 49.03 \\
\bottomrule
\end{tabular}
\label{tab_fid}
\end{table}

To evaluate the visual fidelity of our synthetic weather perturbations relative to real-world corruptions, we conduct experiments on two benchmark datasets: SEN12MS-CR~\cite{ebel2020multisensor} for cloud perturbations and RRSHID~\cite{zhu2025real} for haze perturbations.
For each dataset, we apply our synthetic perturbations to the clean subsets and compute FID against the corresponding real corrupted subsets, which serve as references for natural weather effects. As shown in Table~\ref{tab:fid_comparison}, the FID scores remain below 130 across all severity levels, indicating that the generated perturbations are visually similar to real-world conditions.
For cloud perturbations on SEN12MS-CR, the FID increases consistently with severity, suggesting that lower levels more closely resemble natural cloud coverage, whereas higher levels introduce progressively stronger distortions.
For haze perturbations on RRSHID, the severity range is extended to level 9. The FID decreases up to level 5 and rises thereafter, implying that moderate haze (around level 5) best matches the distribution of real-world haze samples.

\begin{table}[h!]
\centering
\caption{Comparative FID scores for synthetic perturbations against real-world weather datasets across different severity levels. Lower values indicate higher visual similarity.}
\small
\resizebox{0.98\textwidth}{!}{
\begin{tabular}{lccccccccc}
\toprule
\textbf{Dataset} & \textbf{1} & \textbf{2} & \textbf{3} & \textbf{4} & \textbf{5} & \textbf{6} & \textbf{7} & \textbf{8} & \textbf{9} \\
\midrule
SEN12MS-CR (clouds)~\cite{ebel2020multisensor} & 105.56 & 117.27 & 123.72 & 125.70 & 127.03 &  --   &  --   &  --    &  -- \\
RRSHID (haze)~\cite{zhu2025real}       & 107.46 & 100.79 & 90.79 & 80.52 & 73.23 & 73.83 & 88.37 & 128.83 & 190.30 \\
\bottomrule
\end{tabular}
}
\label{tab:fid_comparison}
\end{table}

\section{More Experimental Results}
\subsection{More Implementation Details}
Table~\ref{tab:exp_settings} lists training configurations for different tasks. For the scene classification task, all experiments are conducted on a single NVIDIA RTX 4090 GPU with 24 GB of memory, using the AID dataset, which contains 10,000 high-resolution aerial images. The dataset is split into 8,000 training samples and 2,000 testing samples, with all images uniformly resized to $256 \times 256$ pixels. All models are trained for 100 epochs using the Adam~\cite{kingma2014adam} optimizer with an initial learning rate of 1e-3 and a batch size of 16. A CosineAnnealingWarmRestarts~\cite{loshchilov2016sgdr} scheduler is employed, with the initial cycle length set to 5 epochs.

For the semantic segmentation task, all experiments are conducted on a single NVIDIA A100 GPU with 40 GB of memory. The experiments are implemented using the MMSegmentation~\cite{contributors2020mmsegmentation} framework, with UPerHead used as the primary decoder head and FCNHead as the auxiliary head. All training and evaluation are performed on the Potsdam dataset, where each original image ($6,000 \times 6,000$ pixels) is divided into 144 non-overlapping patches of $512 \times 512$ pixels for model training and inference. The resulting dataset comprises 5,472 image patches, split into 3,456 samples for training and 2,016 for testing. Training is performed for 12 epochs using the AdamW optimizer~\cite{loshchilov2017decoupled}, with an initial learning rate of 6e-5, a batch size of 8, and a weight decay of 0.05.

For the object detection task, all experiments are conducted on a single NVIDIA A100 GPU with 80 GB of memory. The experiments are implemented using the MMRotate framework~\cite{zhou2022mmrotate}, with Oriented R‑CNN employed as the detection head. All images in the DIOR‑R dataset are resized to $800 \times 800$ pixels. The official trainval split is used for training, while the test split is used for both clean evaluation and as the basis for our noise-augmented test set. For fine-tuning, we extracted the last-layer feature map from each model’s pretrained backbone and constructed a four-level feature pyramid via up/down-sampling, which is then fed into a newly initialized Oriented R‑CNN head. The detector is trained for 12 epochs using the AdamW optimizer~\cite{loshchilov2017decoupled}, with an initial learning rate of 1e-5, a batch size of 1, and a learning rate decay of 0.1 applied at epochs 8 and 11.


\begin{table*}[htbp]
\centering
\caption{Training configurations for different tasks.}
\label{tab:exp_settings}
\begin{tabular}{p{0.14\textwidth} p{0.25\textwidth} p{0.25\textwidth} p{0.25\textwidth}}

\hline
 & {Scene Classification} & {Semantic Segmentation} & {Object Detection} \\
\hline
{Dataset} & AID& Potsdam& DIOR-R \\
{Decoder} & Linear & UpperNet & Oriented R-CNN \\
{Optimizer} & Adam & AdamW & AdamW \\
{Epochs} & 100 & 12 & 12 \\
{Lr} & 1e-3 & 6e-5 & 1e-5 \\
{Batch Size} & 16 & 8 & 1 \\
\hline
\end{tabular}
\end{table*}

\subsection{Evaluation Prompts}
For image captioning evaluation, we use GPT-4o-mini to determine for each image whether the predicted captions match the ground truth, with the prompt: \textit{You are trying to tell if a candidate set of captions is describing the same image as a reference set of captions. Candidate set: \{candidate\_statements\}. Reference set: \{target\_statements\}. On a precise scale from 0 to 100, how likely is it that the candidate set is describing the same image as the reference set?(JSON format, with a key ``score'', value between 0 and 100.}

For VQA evaluation, we use GPT-4o-mini to determine for each question whether the answers match ground truth texts, with the prompt: \textit{Question: \{question\}, Ground Truth Answer: \{ground\_truth\}, Predicted Answer: \{predicted answer\}. Does the predicted answer match the ground truth? Answer 1 for match and 0 for not match. Use semantic meaning not exact match. Synonyms are also treated as a match, e.g., pond and swimming pool}.

\subsection{Qualitative Results}

\begin{figure*}[ht!]
    \centering
    \begin{subfigure}[t]{\textwidth}
        \centering
        \begin{subfigure}[t]{0.15\linewidth}
            \centering
            \includegraphics[width=\linewidth]{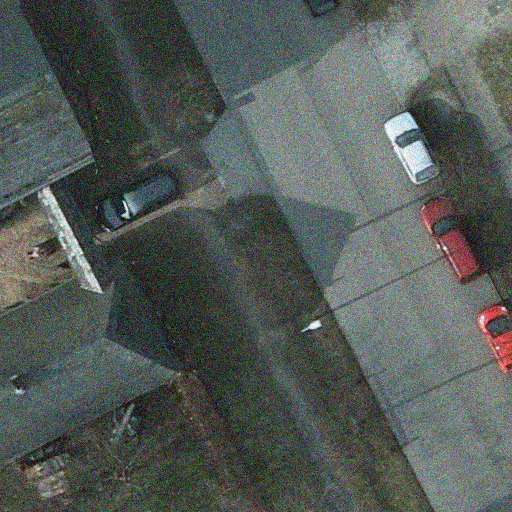}
        \end{subfigure}
        \begin{subfigure}[t]{0.15\linewidth}
            \centering
            \includegraphics[width=\linewidth]{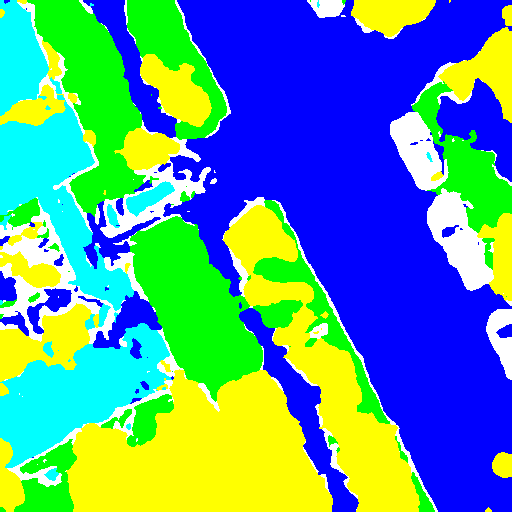}
        \end{subfigure}
        \begin{subfigure}[t]{0.15\linewidth}
            \centering
            \includegraphics[width=\linewidth]{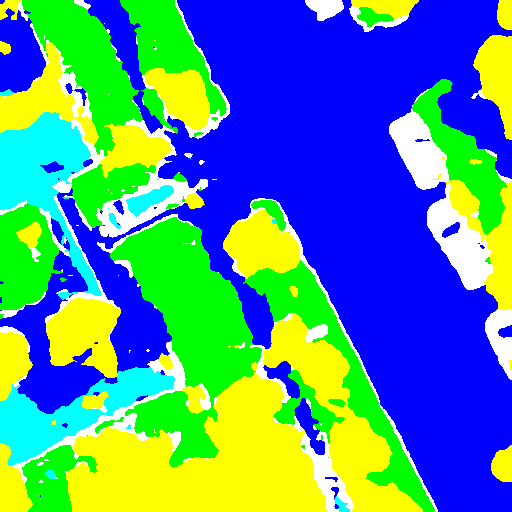}
        \end{subfigure}
        \begin{subfigure}[t]{0.15\linewidth}
            \centering
            \includegraphics[width=\linewidth]{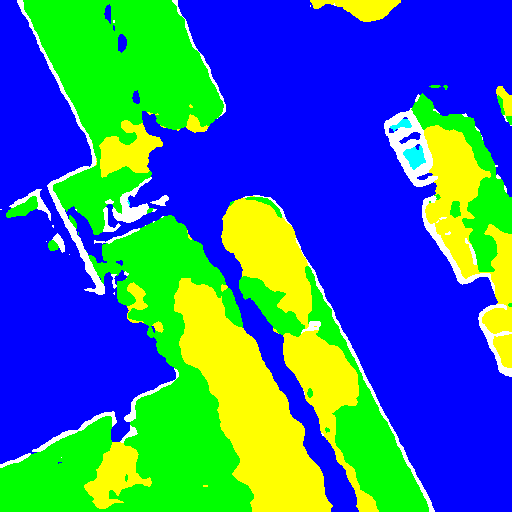}
        \end{subfigure}
        \begin{subfigure}[t]{0.15\linewidth}
            \centering
            \includegraphics[width=\linewidth]{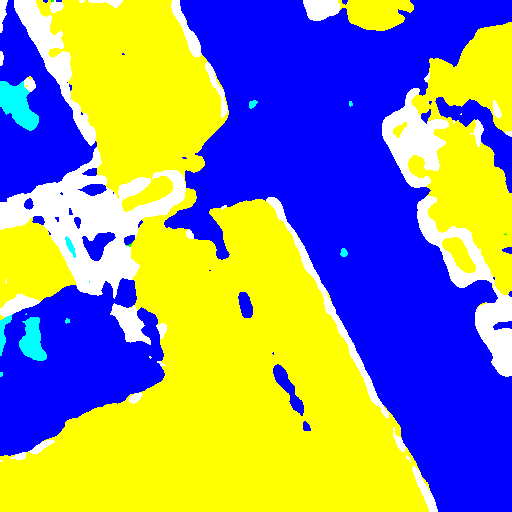}
        \end{subfigure}
        \begin{subfigure}[t]{0.15\linewidth}
            \centering
            \includegraphics[width=\linewidth]{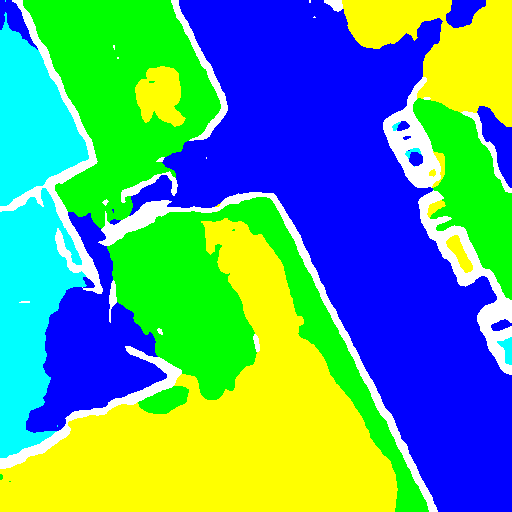}
        \end{subfigure}
        
        \vspace{1pt}
    
        \begin{subfigure}[t]{0.15\linewidth}
            \centering
            \includegraphics[width=\linewidth]{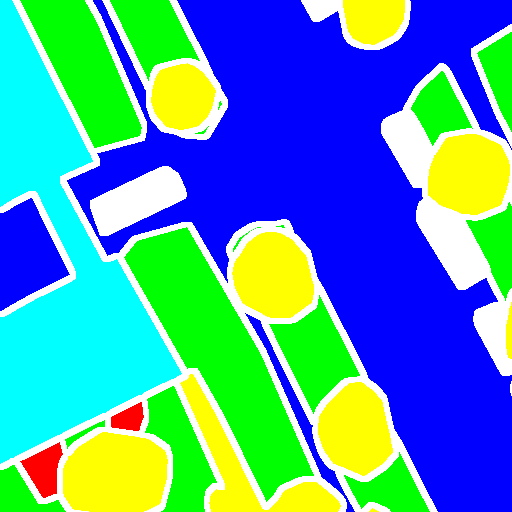}
            \caption*{Ground Truth}
        \end{subfigure}
        \begin{subfigure}[t]{0.15\linewidth}
            \centering
            \includegraphics[width=\linewidth]{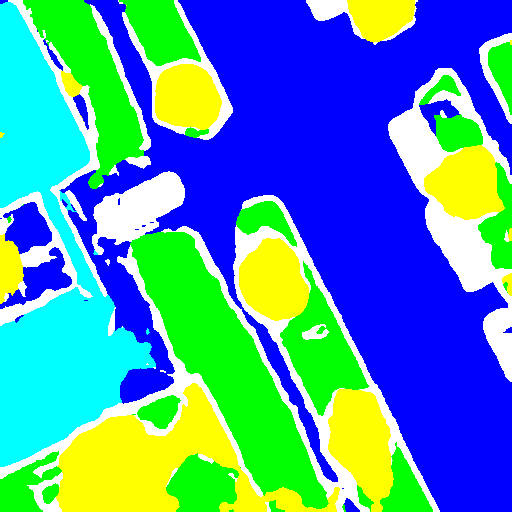}
            \caption*{RemoteCLIP}
        \end{subfigure}
        \begin{subfigure}[t]{0.15\linewidth}
            \centering
            \includegraphics[width=\linewidth]{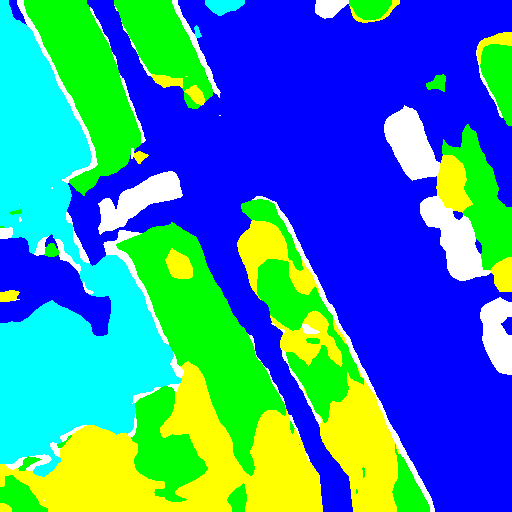}
            \caption*{GeoRSCLIP}
        \end{subfigure}
        \begin{subfigure}[t]{0.15\linewidth}
            \centering
            \includegraphics[width=\linewidth]{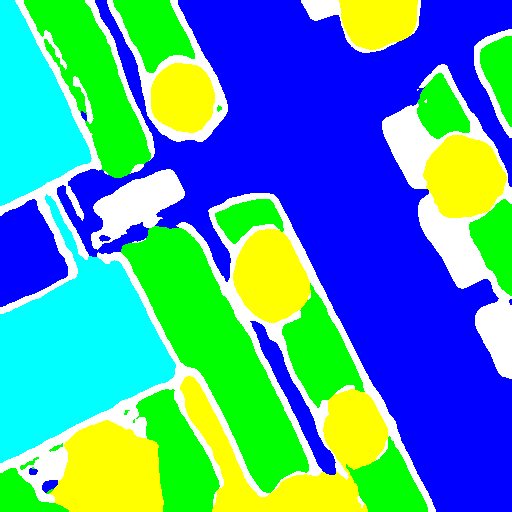}
            \caption*{RVSA}
        \end{subfigure}
        \begin{subfigure}[t]{0.15\linewidth}
            \centering
            \includegraphics[width=\linewidth]{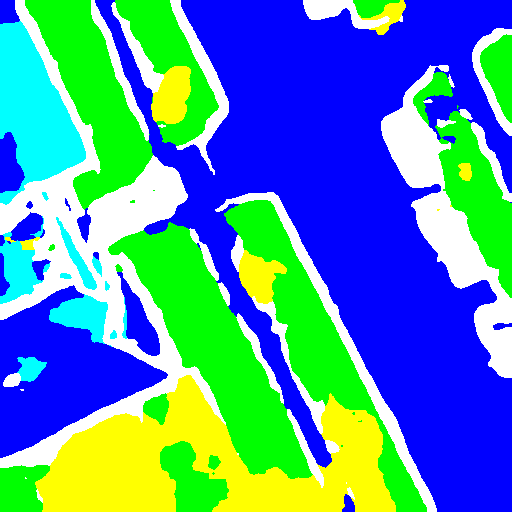}
            \caption*{SatMAE++}
        \end{subfigure}
        \begin{subfigure}[t]{0.15\linewidth}
            \centering
            \includegraphics[width=\linewidth]{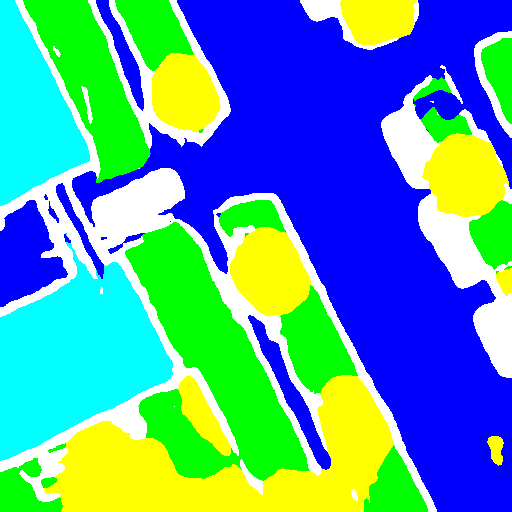}
            \caption*{ScaleMAE}
        \end{subfigure}
    \caption{}
    \end{subfigure}
    \begin{subfigure}[t]{\textwidth}
        \centering
        \begin{subfigure}[t]{0.15\linewidth}
            \centering
            \includegraphics[width=\linewidth]{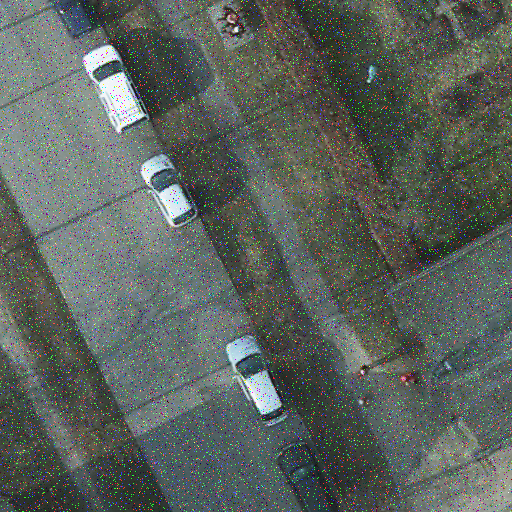}
        \end{subfigure}
        \begin{subfigure}[t]{0.15\linewidth}
            \centering
            \includegraphics[width=\linewidth]{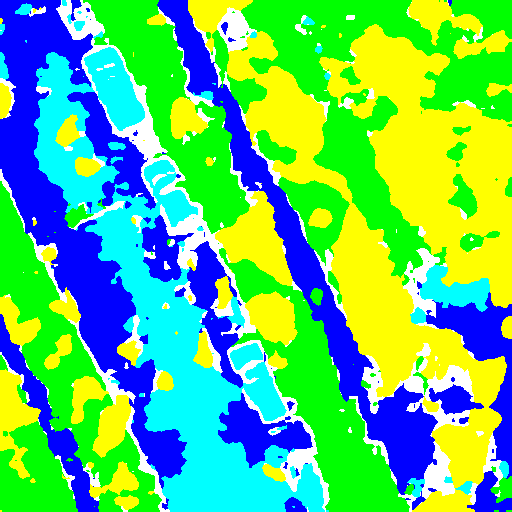}
        \end{subfigure}
        \begin{subfigure}[t]{0.15\linewidth}
            \centering
            \includegraphics[width=\linewidth]{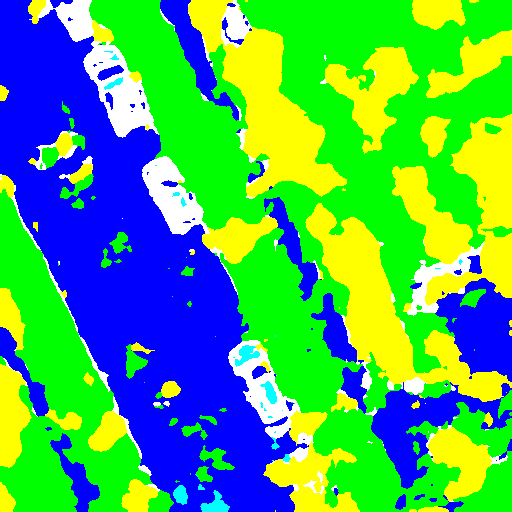}
        \end{subfigure}
        \begin{subfigure}[t]{0.15\linewidth}
            \centering
            \includegraphics[width=\linewidth]{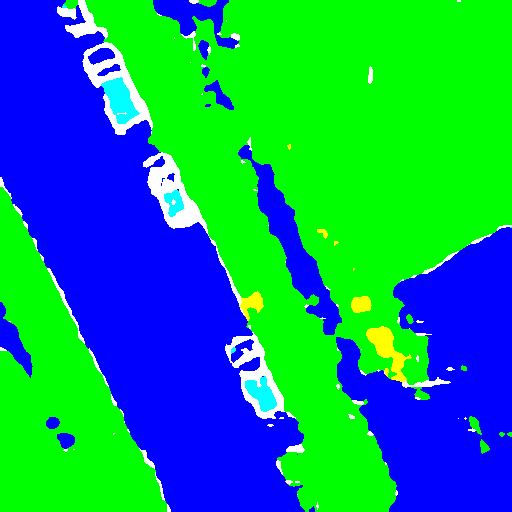}
        \end{subfigure}
        \begin{subfigure}[t]{0.15\linewidth}
            \centering
            \includegraphics[width=\linewidth]{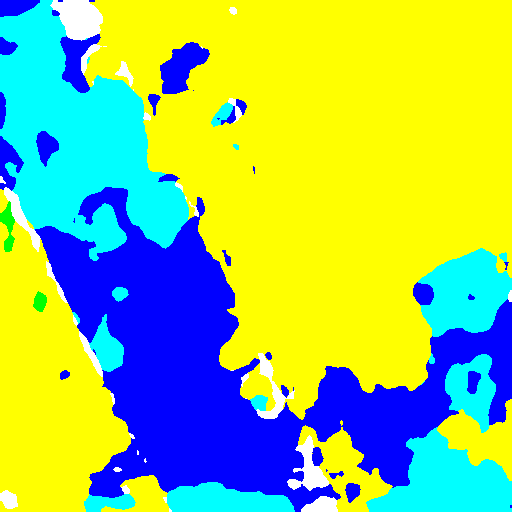}
        \end{subfigure}
        \begin{subfigure}[t]{0.15\linewidth}
            \centering
            \includegraphics[width=\linewidth]{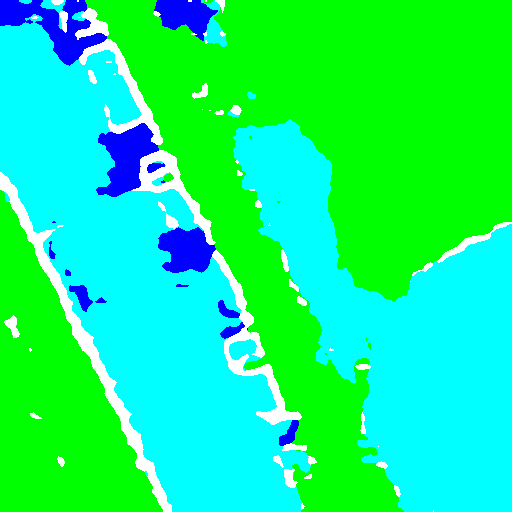}
        \end{subfigure}
        
        \vspace{1pt}
    
        \begin{subfigure}[t]{0.15\linewidth}
            \centering
            \includegraphics[width=\linewidth]{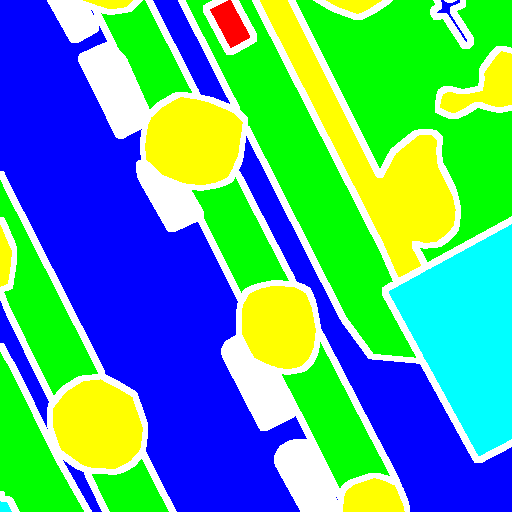}
            \caption*{Ground Truth}
        \end{subfigure}
        \begin{subfigure}[t]{0.15\linewidth}
            \centering
            \includegraphics[width=\linewidth]{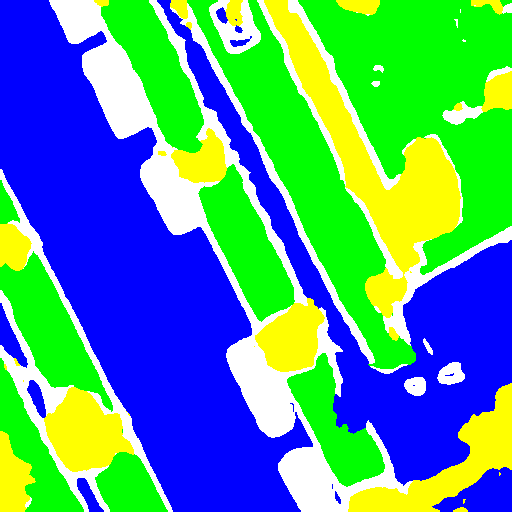}
            \caption*{RemoteCLIP}
        \end{subfigure}
        \begin{subfigure}[t]{0.15\linewidth}
            \centering
            \includegraphics[width=\linewidth]{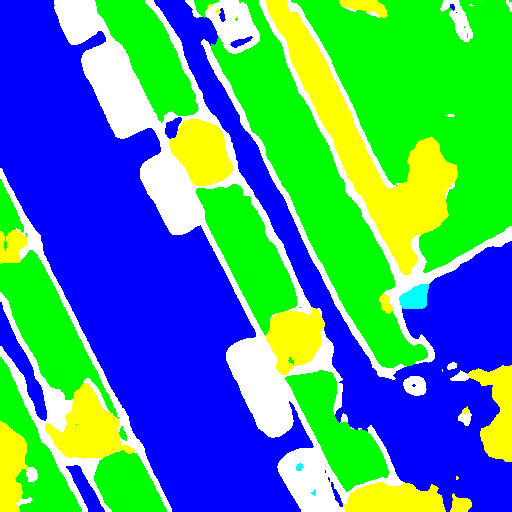}
            \caption*{GeoRSCLIP}
        \end{subfigure}
        \begin{subfigure}[t]{0.15\linewidth}
            \centering
            \includegraphics[width=\linewidth]{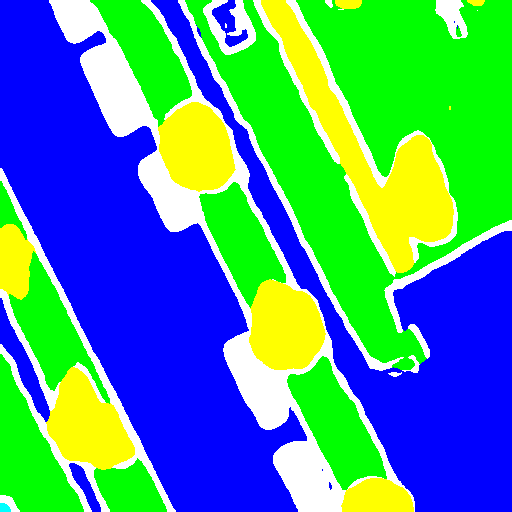}
            \caption*{RVSA}
        \end{subfigure}
        \begin{subfigure}[t]{0.15\linewidth}
            \centering
            \includegraphics[width=\linewidth]{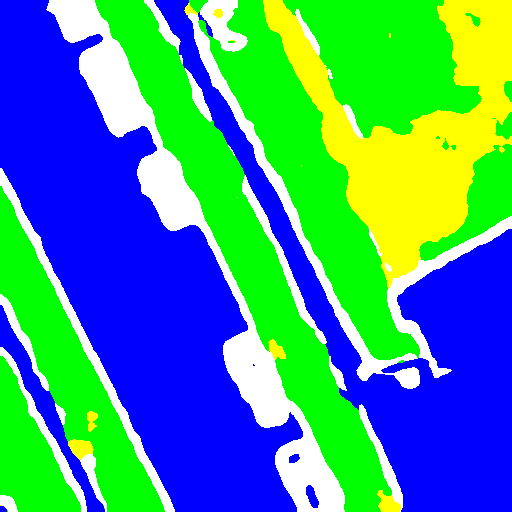}
            \caption*{SatMAE++}
        \end{subfigure}
        \begin{subfigure}[t]{0.15\linewidth}
            \centering
            \includegraphics[width=\linewidth]{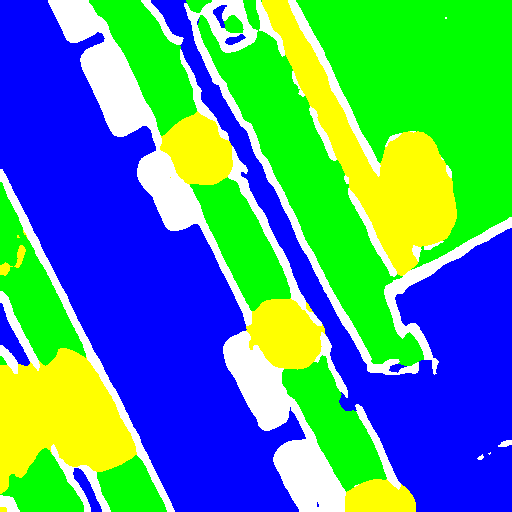}
            \caption*{ScaleMAE}
        \end{subfigure}
    \caption{}
    \end{subfigure}

    \caption{Selected semantic segmentation examples under (a) Gaussian noise (severity 5) and (b) Salt-and-pepper noise (severity 5). For each example, the top row shows the corrupted image and the segmentation results of different models under this corruption, and the bottom row shows the ground truth and segmentation results on clean images.}
    \label{fig:seg_visualise}
\end{figure*}


\begin{figure*}[ht!]
    \centering
    \begin{subfigure}[t]{\textwidth}
        \centering
        \begin{subfigure}[t]{0.15\linewidth}
            \centering
            \includegraphics[width=\linewidth]{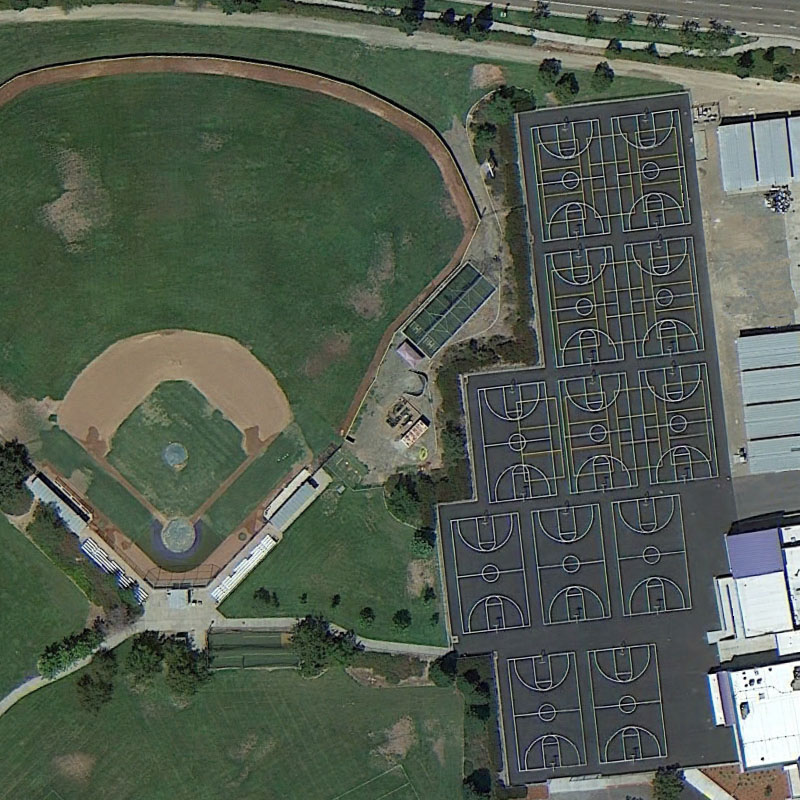}
        \end{subfigure}
        \begin{subfigure}[t]{0.15\linewidth}
            \centering
            \includegraphics[width=\linewidth]{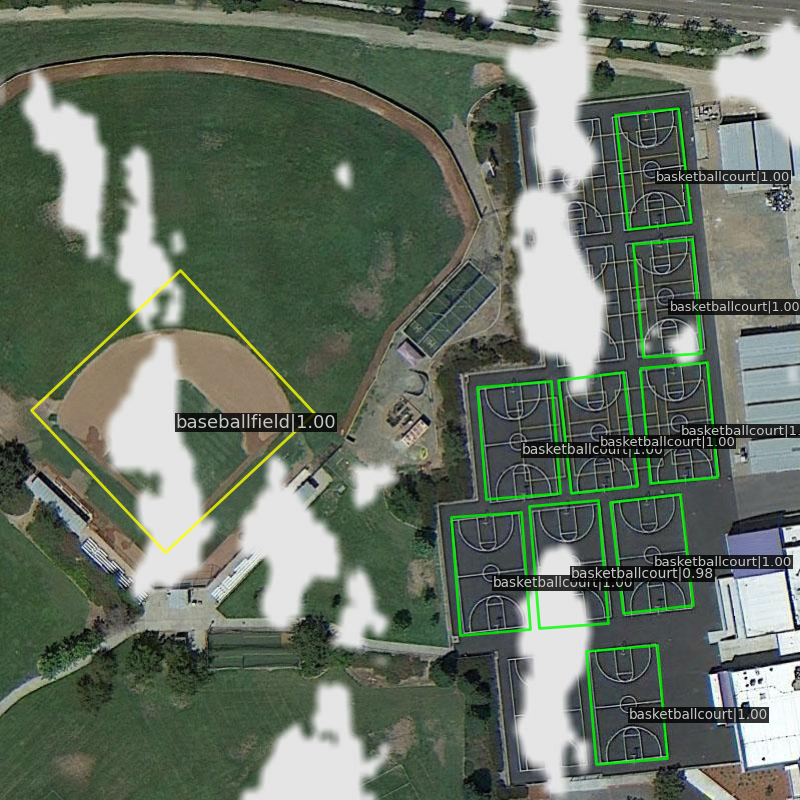}
        \end{subfigure}
        \begin{subfigure}[t]{0.15\linewidth}
            \centering
            \includegraphics[width=\linewidth]{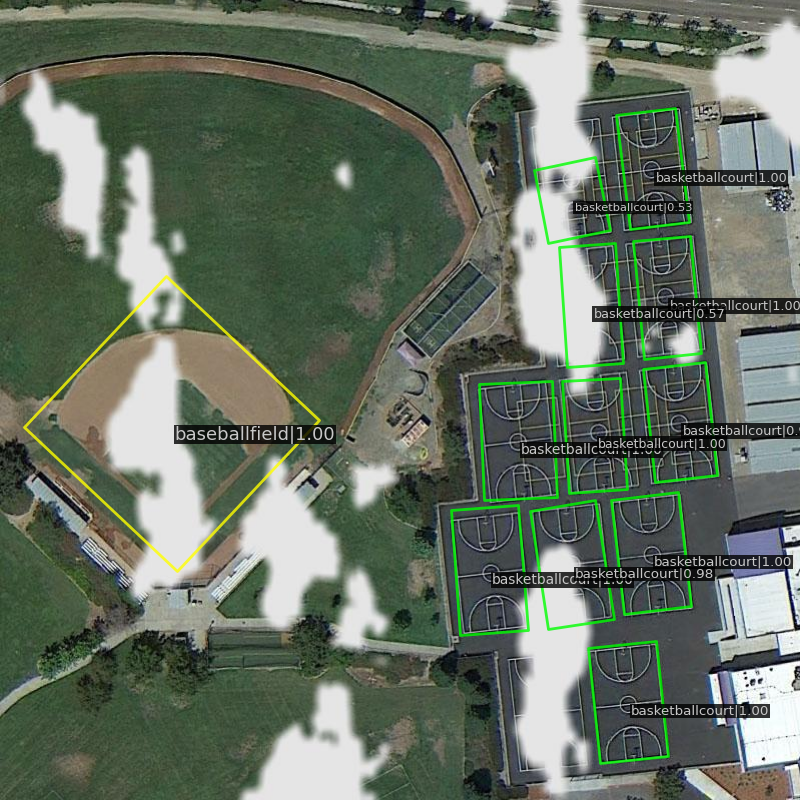}
        \end{subfigure}
        \begin{subfigure}[t]{0.15\linewidth}
            \centering
            \includegraphics[width=\linewidth]{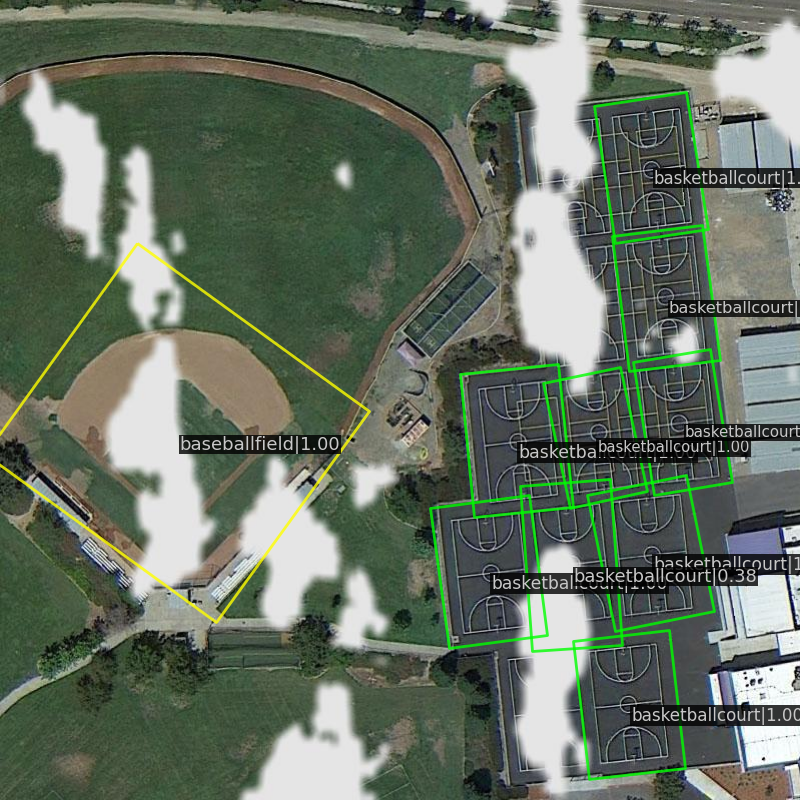}
        \end{subfigure}
        \begin{subfigure}[t]{0.15\linewidth}
            \centering
            \includegraphics[width=\linewidth]{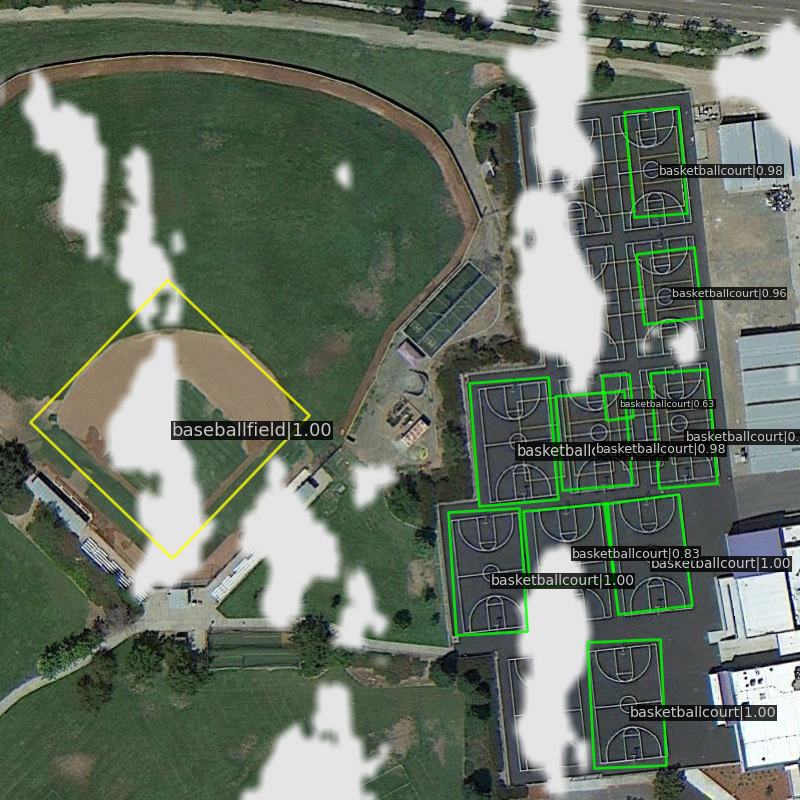}
        \end{subfigure}
        \begin{subfigure}[t]{0.15\linewidth}
            \centering
            \includegraphics[width=\linewidth]{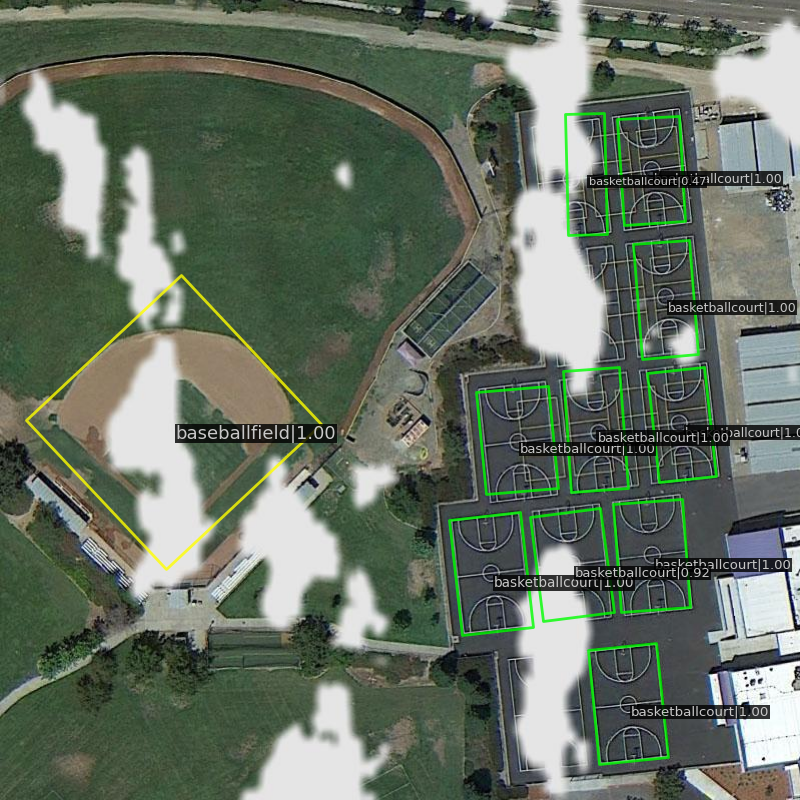}
        \end{subfigure}
        
        \vspace{1pt}

        \begin{subfigure}[t]{0.15\linewidth}
            \centering
            \includegraphics[width=\linewidth]{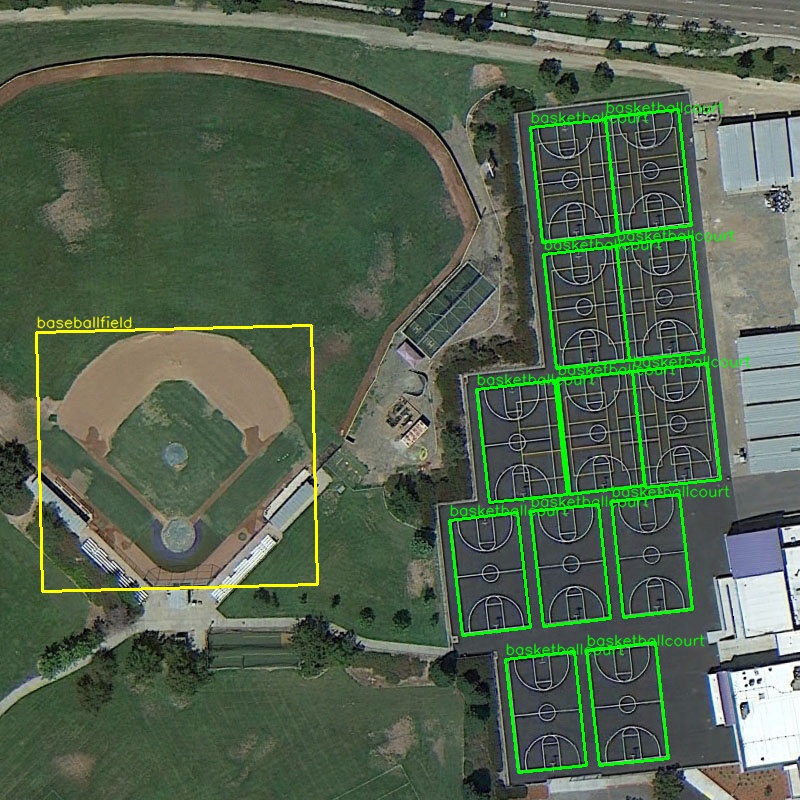}
            \caption*{Ground Truth}
        \end{subfigure}
        \begin{subfigure}[t]{0.15\linewidth}
            \centering
            \includegraphics[width=\linewidth]{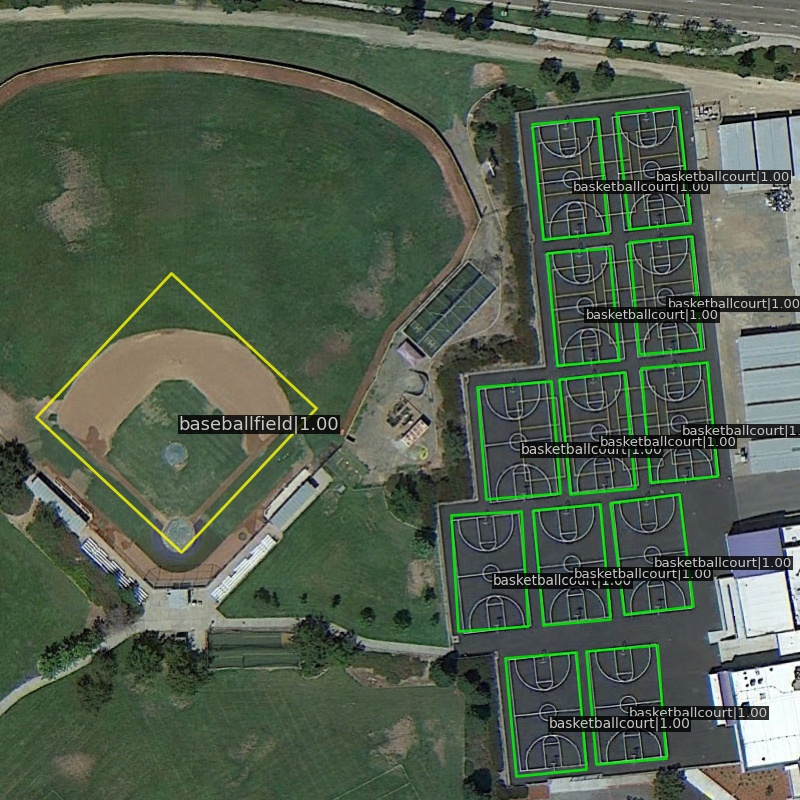}
            \caption*{RemoteCLIP}
        \end{subfigure}
        \begin{subfigure}[t]{0.15\linewidth}
            \centering
            \includegraphics[width=\linewidth]{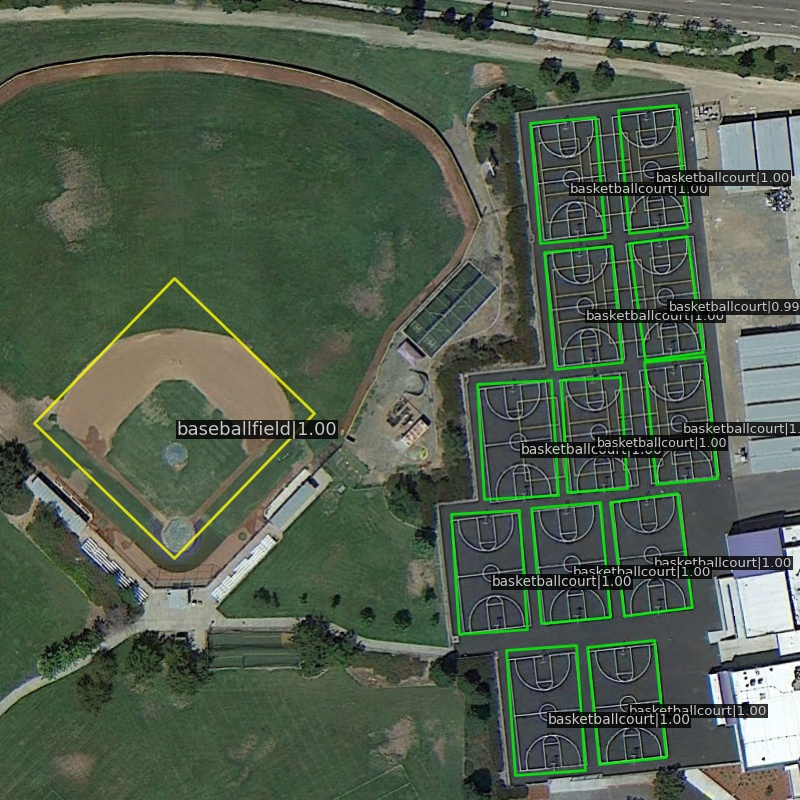}
            \caption*{GeoRSCLIP}
        \end{subfigure}
        \begin{subfigure}[t]{0.15\linewidth}
            \centering
            \includegraphics[width=\linewidth]{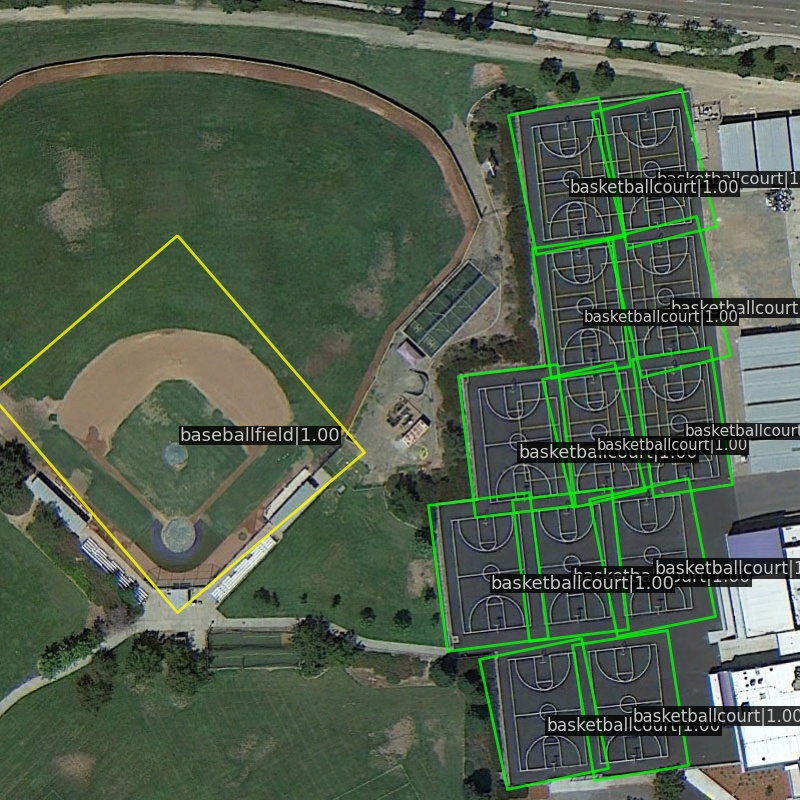}
            \caption*{RVSA}
        \end{subfigure}
        \begin{subfigure}[t]{0.15\linewidth}
            \centering
            \includegraphics[width=\linewidth]{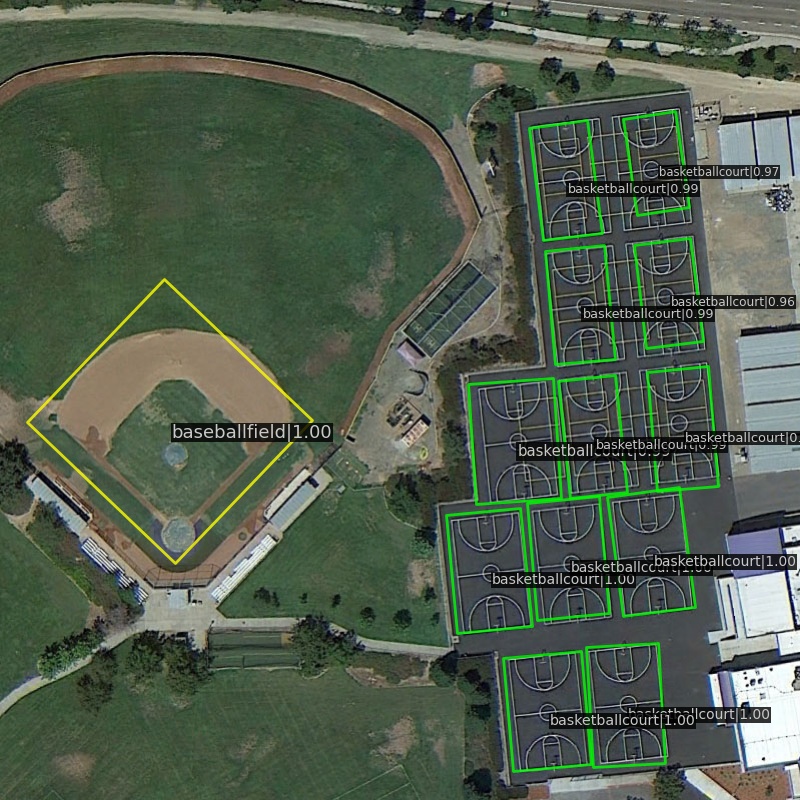}
            \caption*{SatMAE++}
        \end{subfigure}
        \begin{subfigure}[t]{0.15\linewidth}
            \centering
                \includegraphics[width=\linewidth]{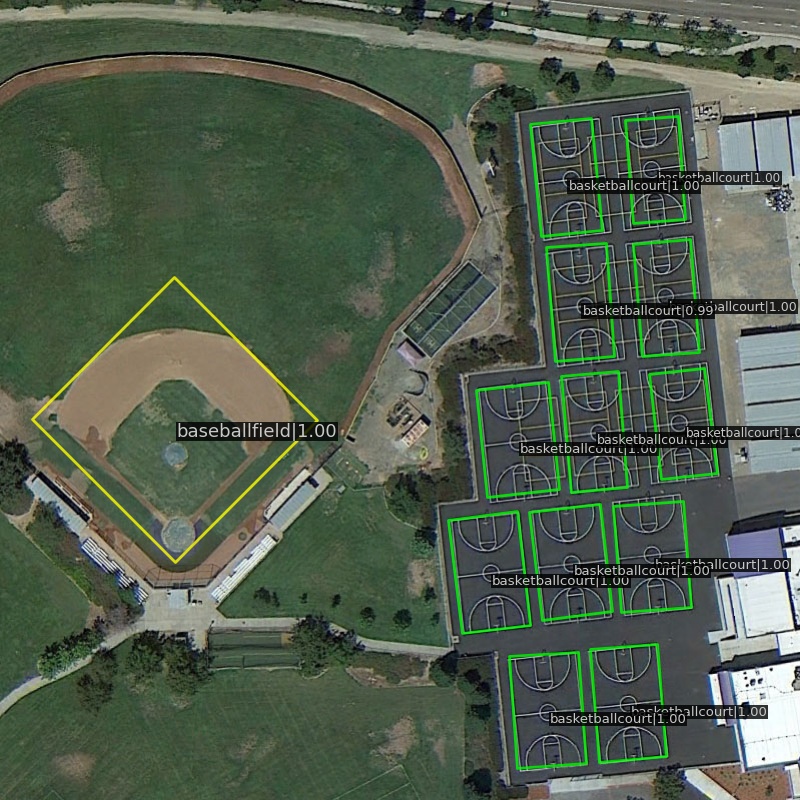}
            \caption*{ScaleMAE}
        \end{subfigure}
        \caption{}
    \end{subfigure}
    
    \begin{subfigure}[t]{\textwidth}
        \centering
        \begin{subfigure}[t]{0.15\linewidth}
            \centering
            \includegraphics[width=\linewidth]{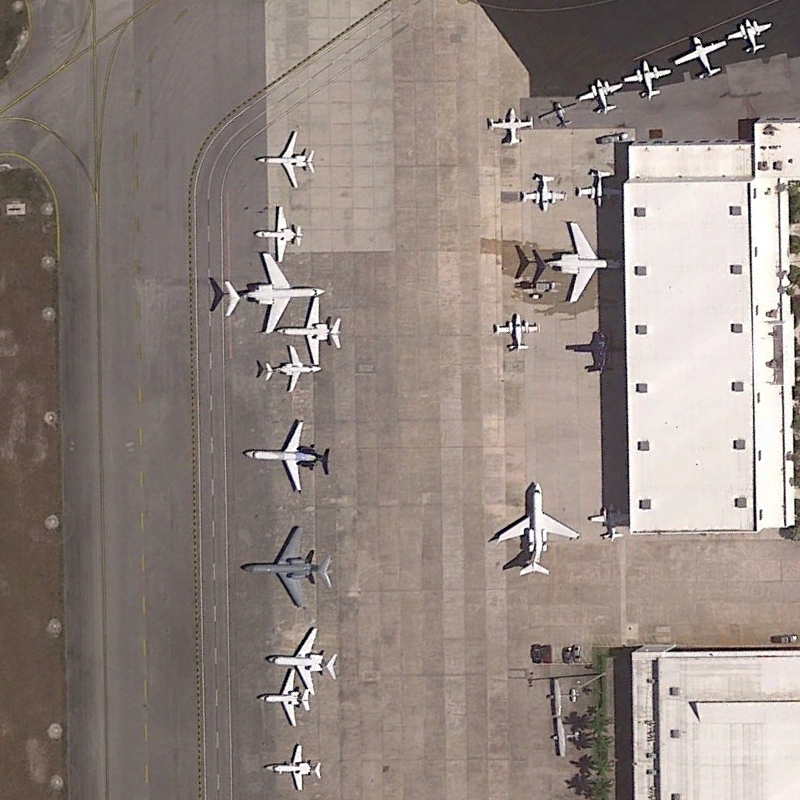}
        \end{subfigure}
        \begin{subfigure}[t]{0.15\linewidth}
            \centering
            \includegraphics[width=\linewidth]{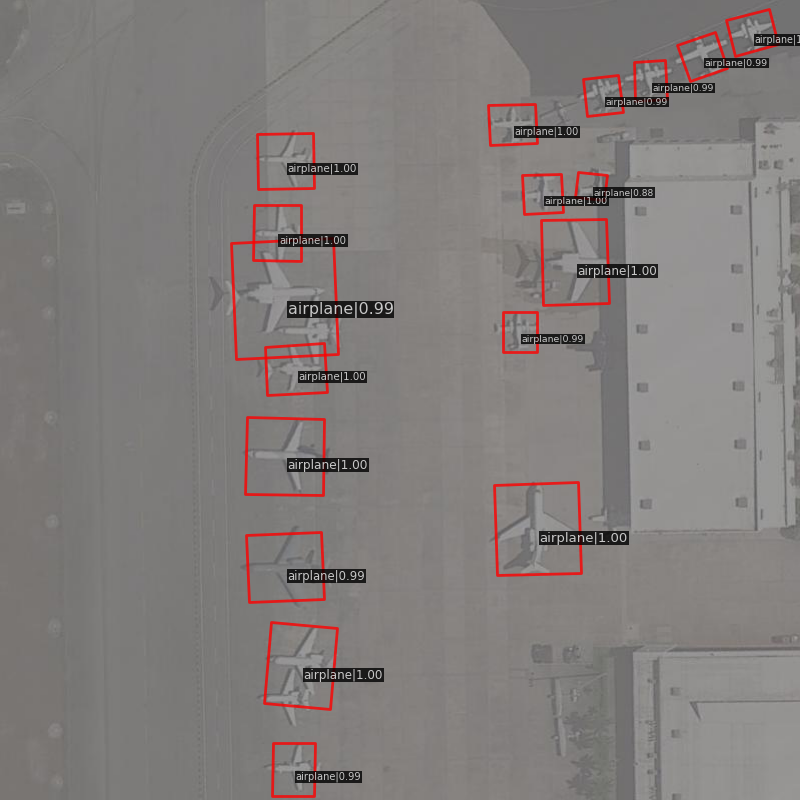}
        \end{subfigure}
        \begin{subfigure}[t]{0.15\linewidth}
            \centering
            \includegraphics[width=\linewidth]{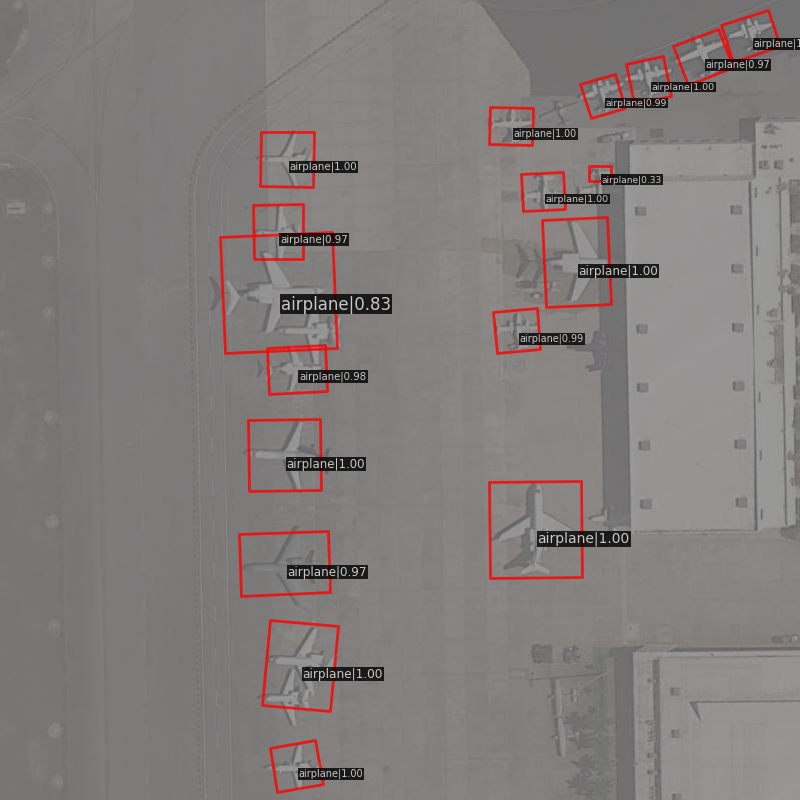}
        \end{subfigure}
        \begin{subfigure}[t]{0.15\linewidth}
            \centering
            \includegraphics[width=\linewidth]{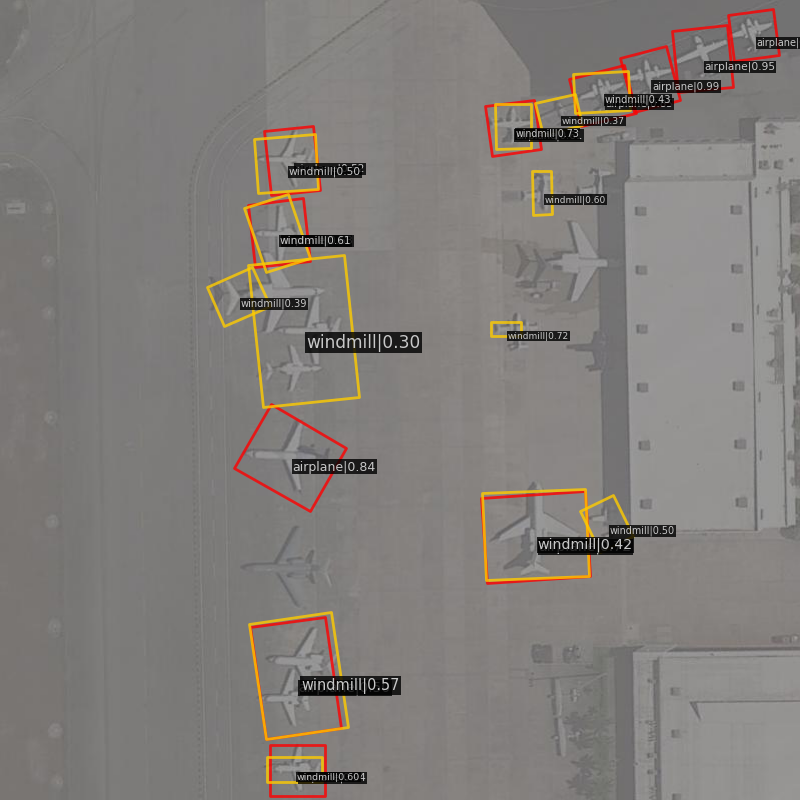}
        \end{subfigure}
        \begin{subfigure}[t]{0.15\linewidth}
            \centering
            \includegraphics[width=\linewidth]{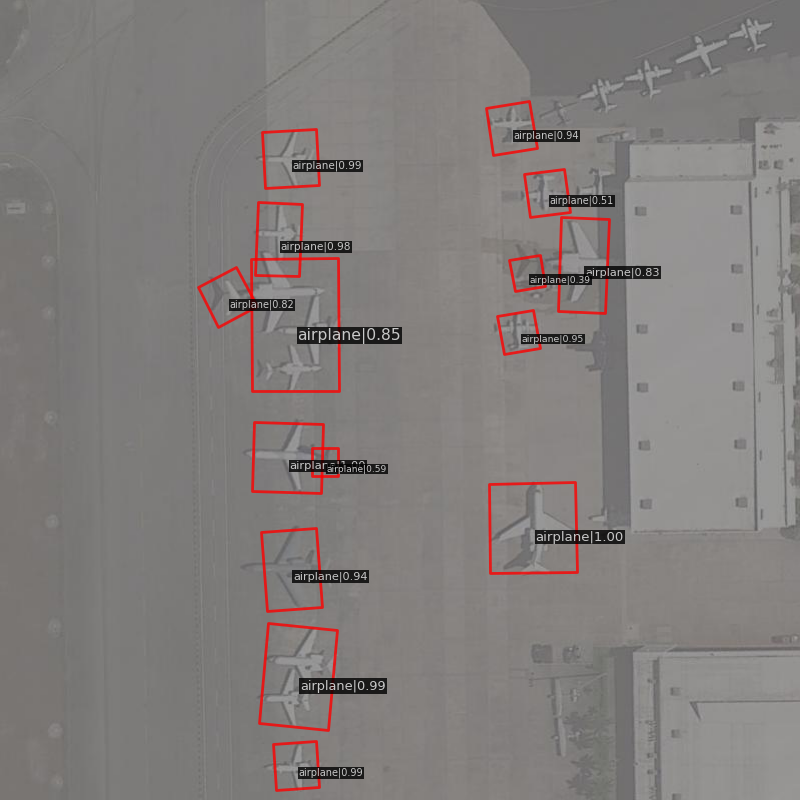}
        \end{subfigure}
        \begin{subfigure}[t]{0.15\linewidth}
            \centering
            \includegraphics[width=\linewidth]{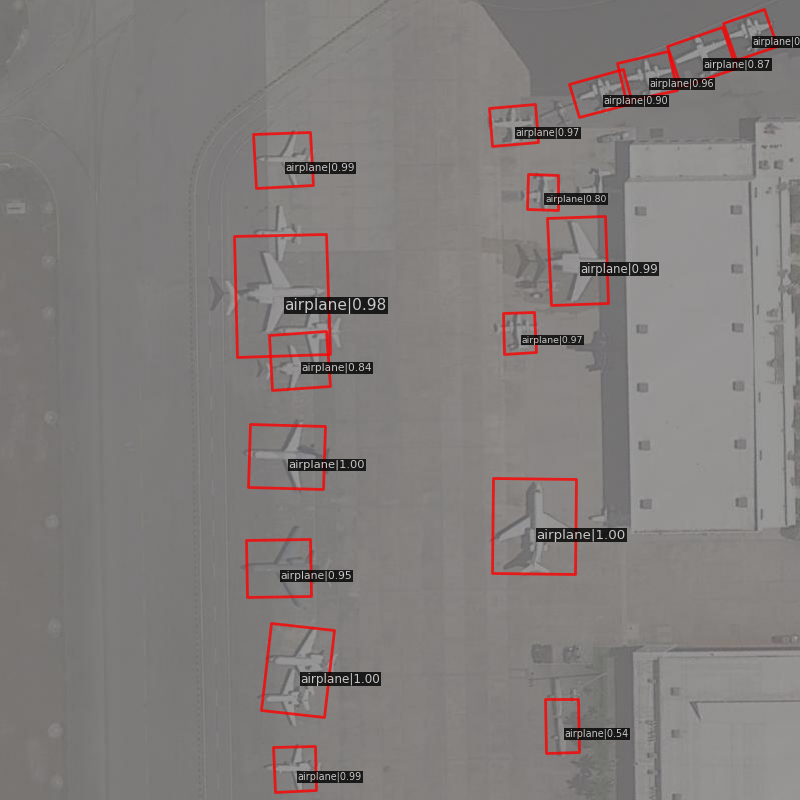}
        \end{subfigure}
        
        \vspace{1pt}

        \begin{subfigure}[t]{0.15\linewidth}
            \centering
            \includegraphics[width=\linewidth]{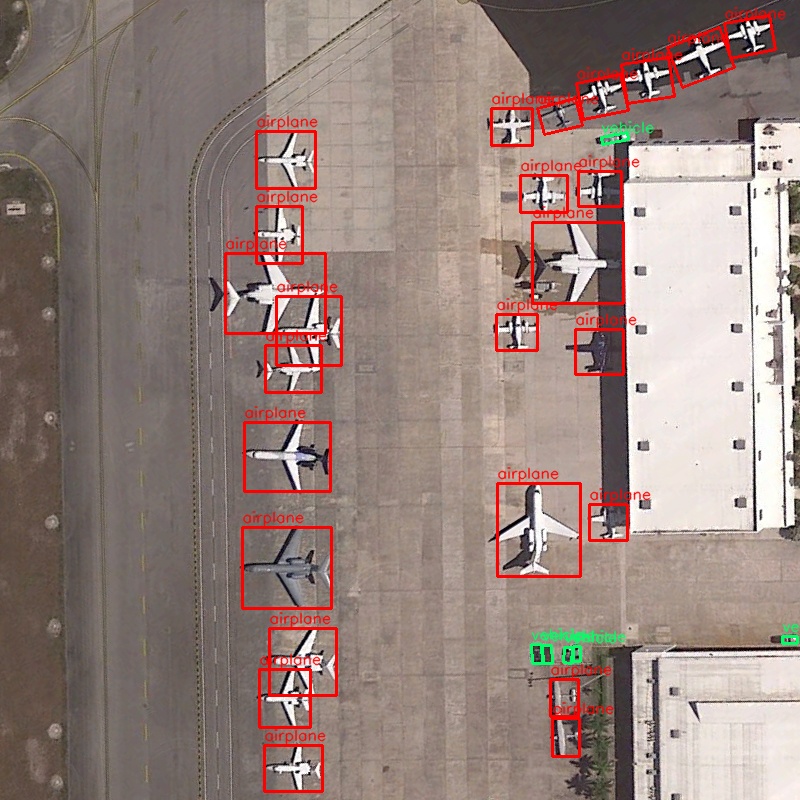}
            \caption*{Ground Truth}
        \end{subfigure}
        \begin{subfigure}[t]{0.15\linewidth}
            \centering
            \includegraphics[width=\linewidth]{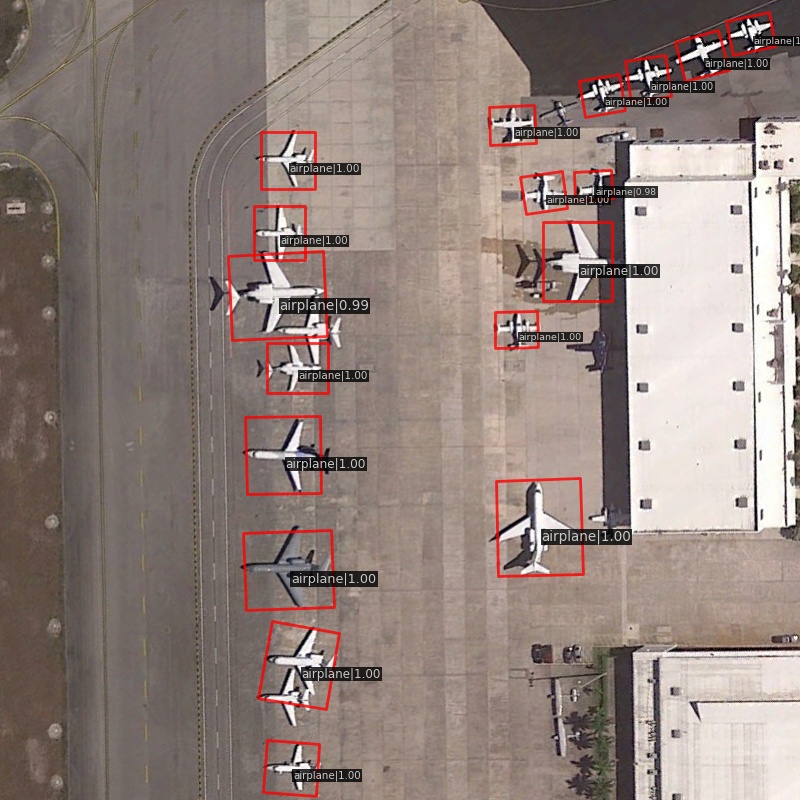}
            \caption*{RemoteCLIP}
        \end{subfigure}
        \begin{subfigure}[t]{0.15\linewidth}
            \centering
            \includegraphics[width=\linewidth]{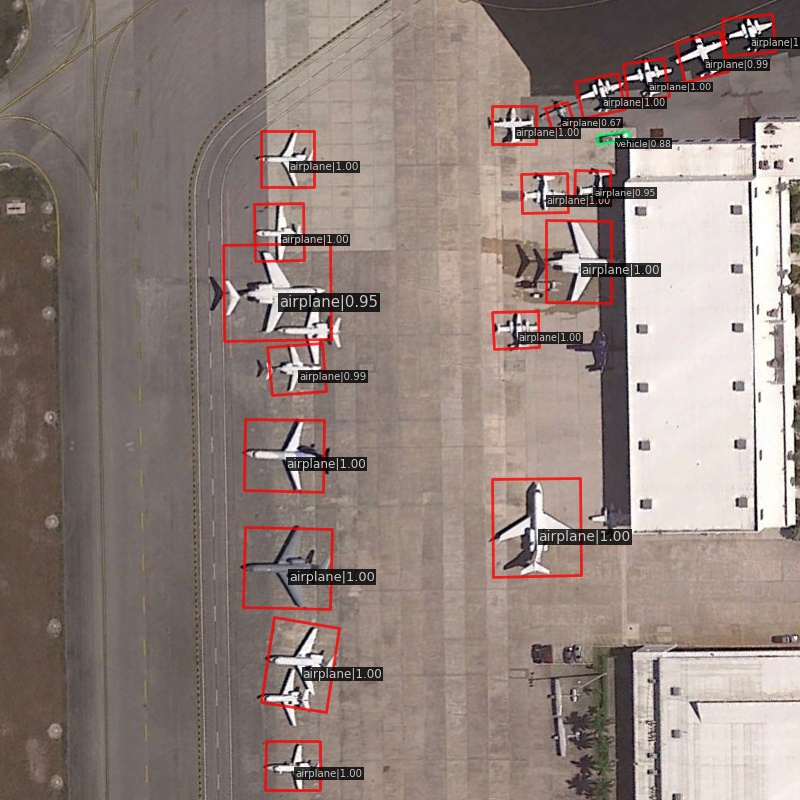}
            \caption*{GeoRSCLIP}
        \end{subfigure}
        \begin{subfigure}[t]{0.15\linewidth}
            \centering
            \includegraphics[width=\linewidth]{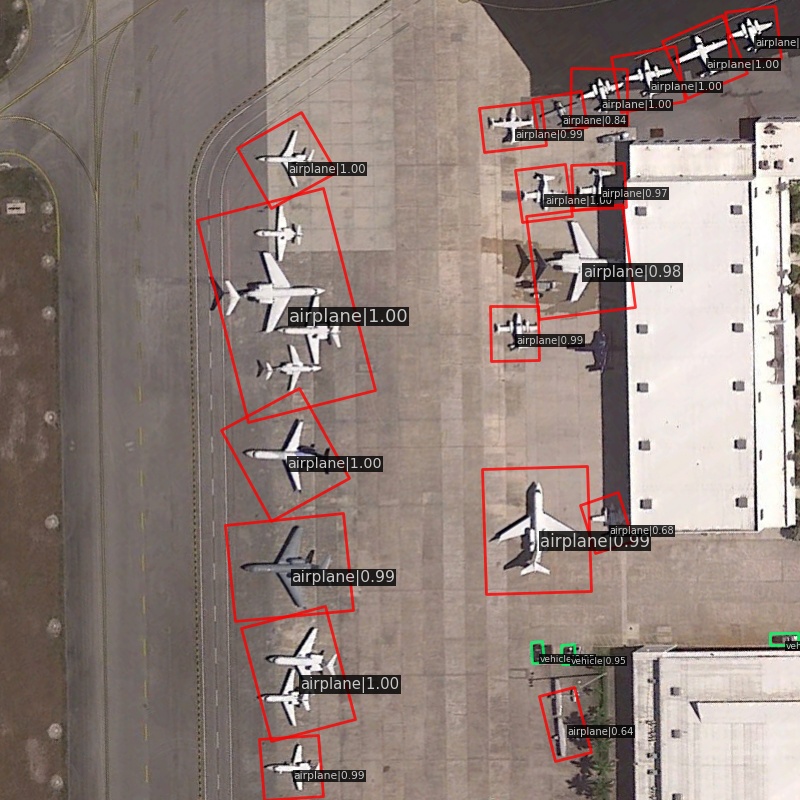}
            \caption*{RVSA}
        \end{subfigure}
        \begin{subfigure}[t]{0.15\linewidth}
            \centering
            \includegraphics[width=\linewidth]{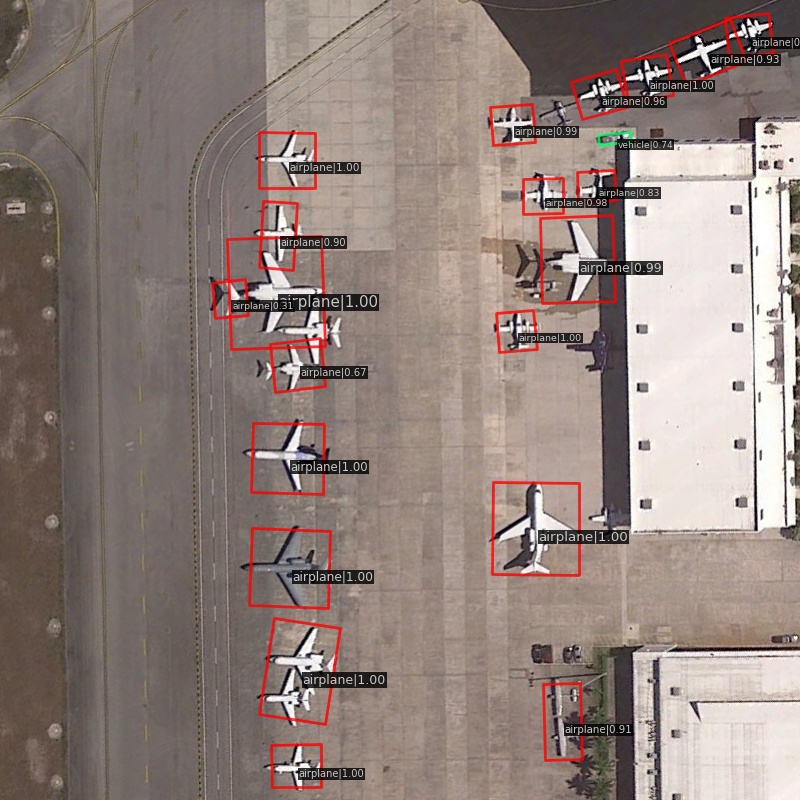}
            \caption*{SatMAE++}
        \end{subfigure}
        \begin{subfigure}[t]{0.15\linewidth}
            \centering
            \includegraphics[width=\linewidth]{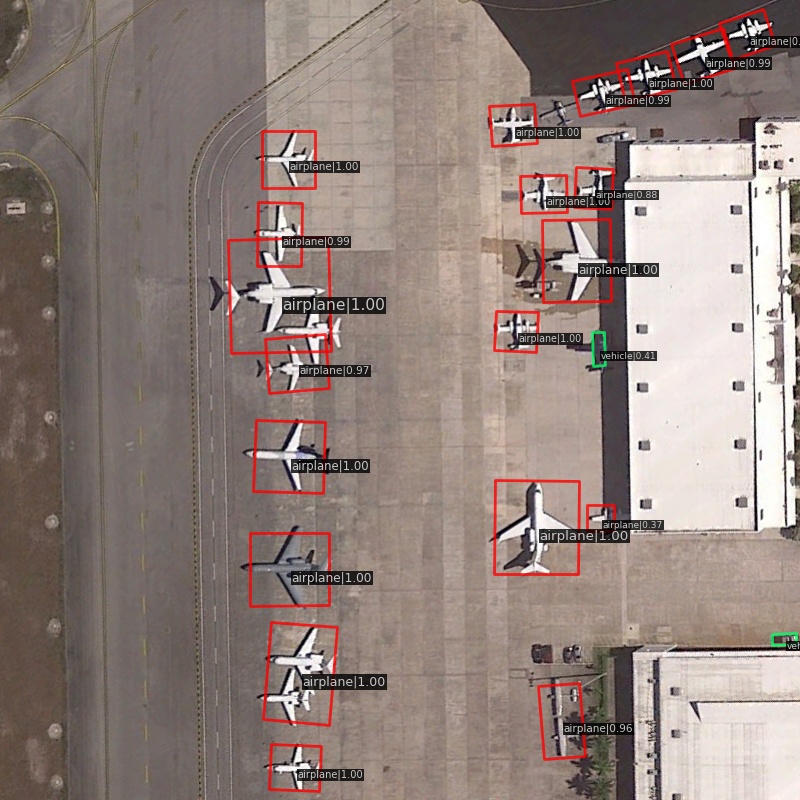}
            \caption*{ScaleMAE}
        \end{subfigure}
        \caption{}
    \end{subfigure}

    \caption{Selected object detection examples under (a) cloud (severity 5) and (b) brightness‑and‑contrast corruption (severity 5). For each example, the top row shows the corrupted image and the detection results of different models under this corruption, and the bottom row shows the ground truth and detection results on clean images.}
    \label{fig:det_visualise}
\end{figure*}


\begin{figure}[ht!]
    \centering
  \begin{minipage}[t]{0.18\textwidth}
    \vspace{0pt}
    \includegraphics[width=\linewidth]{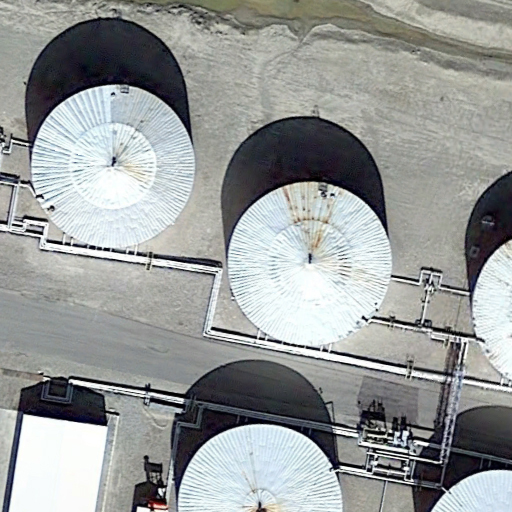}
    \smallskip
    \includegraphics[width=\linewidth]{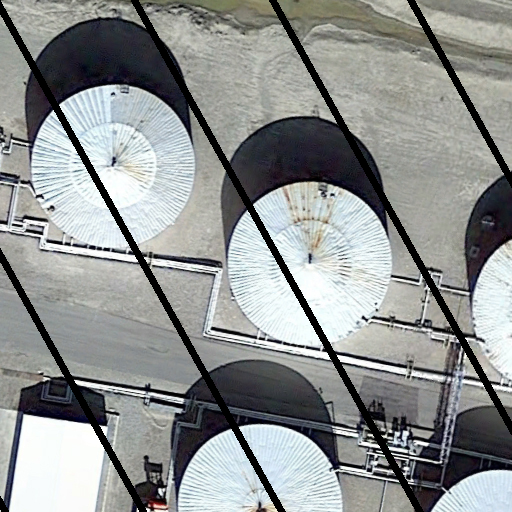}
    \smallskip
    \centering
    (a) 
  \end{minipage}%
    \hspace{0.02\textwidth}
    \begin{minipage}[t]{0.77\textwidth}
        \vspace{0pt}
        \raggedright
        \small
        Top image: {05925\_0000.png}\\
        Bottom image: \textbf{Corruption type:} Data Gaps, \textbf{Severity:} 5.\\[6pt]
        \textbf{GT:} The image, sourced from GoogleEarth, shows a facility featuring large \textcolor{green}{storage tanks} with distinctive \textcolor{green}{white} circular structures against a \textcolor{green}{grey background}. There are \textcolor{green}{three} visible tanks, arranged from top to bottom within the frame.\\[4pt]
        \textbf{RS-LLaVA:} A factory features {numerous} \textcolor{green}{storage tanks}, with some being \textcolor{green}{white} and others having a silver appearance.\\[4pt]
        \textbf{VHM:} The image shows an aerial view of a large oil depot. There are \textcolor{red}{eight} large \textcolor{green}{white storage tanks} arranged in a square. Each tank has a black ring around the top. There is a road between the tanks and a white building to the left of the tanks. \\[4pt]
        \textbf{Falcon:} the oil tanker is equipped with \textcolor{red}{a large number} of oil tanks in the dark. yellow and white neatly placed in the oil \textcolor{green}{storage tank} on the ship. \textcolor{green}{three} large storage tanks are in a piece of bareland. \\[4pt]
  \end{minipage}
    
  \vspace{10pt}

  \begin{minipage}[t]{0.18\textwidth}
    \vspace{0pt}
    \includegraphics[width=\linewidth]{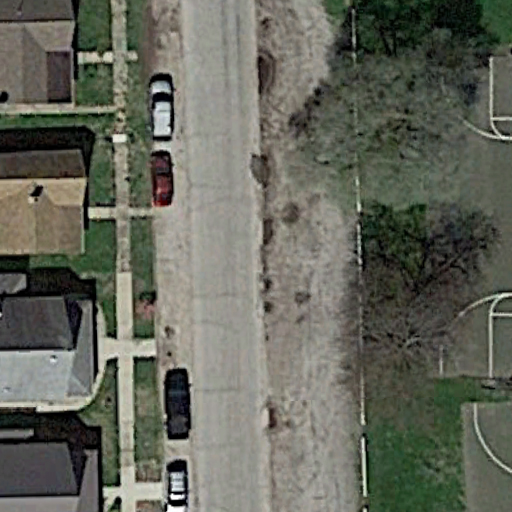}
    \smallskip
    \includegraphics[width=\linewidth]{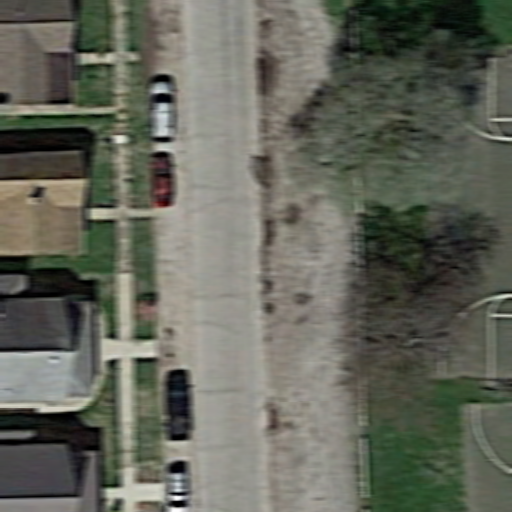}
    \smallskip
    \centering
    (b) 
  \end{minipage}%
    \hspace{0.02\textwidth}
    \begin{minipage}[t]{0.77\textwidth}
        \vspace{0pt}
        \raggedright
        \small
        Top image: {05870\_0000.png}\\
        Bottom image: \textbf{Corruption type:} Motion Blur, \textbf{Severity:} 5.\\[6pt]
        \textbf{GT:} Aerial view of \textcolor{green}{a residential area} from Google Earth showing a portion of a \textcolor{green}{street} with \textcolor{green}{vehicles} lined up. The capture shows \textcolor{green}{four} parked on the side of the street adjacent to houses. The \textcolor{green}{houses} have various roof colors and lawn areas.\\[4pt]
        \textbf{RS-LLaVA:} A \textcolor{red}{basketball court} sits on a \textcolor{green}{lawn}, surrounded by \textcolor{green}{buildings and trees}, with a \textcolor{green}{road} nearby. \\[4pt]
        \textbf{VHM:} The image shows a \textcolor{green}{residential area} with a \textcolor{green}{tree-lined street}. There are \textcolor{green}{houses} on \textcolor{red}{both sides} of the street. The houses are mostly made of brick and have \textcolor{green}{different colors}. The street is made of asphalt and has \textcolor{green}{cars parked} on it. There is also a basketball court on the right side of the image. \\[4pt]
        \textbf{Falcon:} a \textcolor{green}{road} with \textcolor{red}{two} cars and a basketball field is next to a piece of \textcolor{green}{green meadow} and \textcolor{green}{several buildings}. two cars are \textcolor{red}{running} on the wide road and there are some buildings on one side of the road. \\[4pt]
    \end{minipage}
    
    \caption{Selected image captioning examples. We highlight correct information in green and incorrect information in red.}
    \label{fig:cap_visualise}

\end{figure}


\begin{figure}[ht!]
    \centering
  \begin{minipage}[t]{0.18\textwidth}
    \vspace{0pt}
    \includegraphics[width=\linewidth]{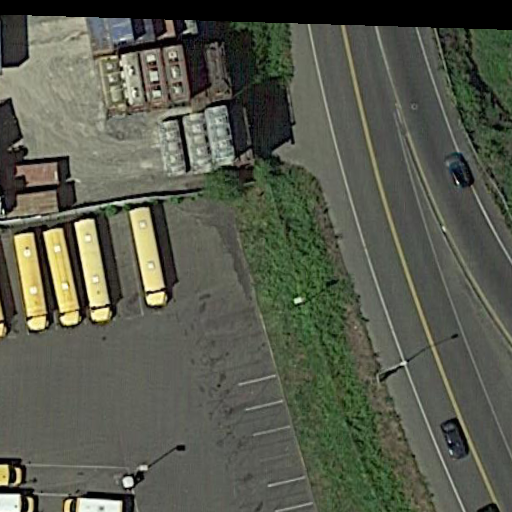}
    \smallskip
    \includegraphics[width=\linewidth]{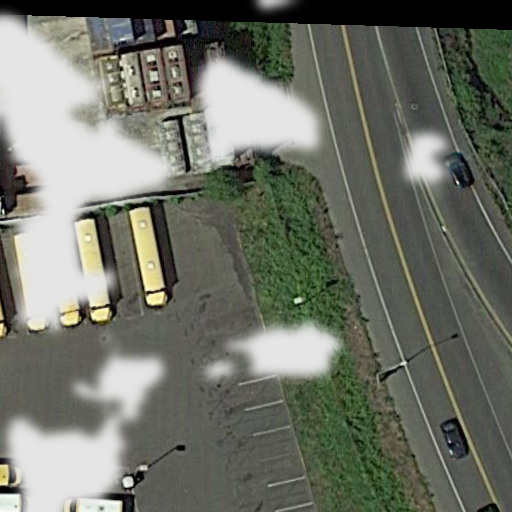}
    \smallskip
    \centering
    (a)
  \end{minipage}%
    \hspace{0.02\textwidth}
    \begin{minipage}[t]{0.77\textwidth}
        \vspace{0pt}
        \raggedright
        \small
        Top image: \textbf{P0003\_0002.png}\\
        Bottom image: \textbf{Corruption type:} Cloud, \textbf{Severity:} 5.\\[6pt]
        \textit{What color are the large vehicles seen in the image?}\\
        \textbf{GT}: \textcolor{green}{Yellow}, 
        \textbf{GeoChat}: \textcolor{red}{Teal}, 
        \textbf{SkySenseGPT}: \textcolor{red}{teal}, 
        \textbf{VHM}: \textcolor{green}{yellow}.\\[4PT]
        \textit{How many small vehicles are visible in the image?}\\
        \textbf{GT}: \textcolor{green}{2}, 
        \textbf{GeoChat}: \textcolor{red}{11}, 
        \textbf{SkySenseGPT}: \textcolor{green}{2},  
        \textbf{VHM}: \textcolor{red}{3}.\\[4PT]
        \textit{Is there a vehicle located at the top-most position in the provided image?}\\
        \textbf{GT}: \textcolor{green}{Yes}, 
        \textbf{GeoChat}: \textcolor{green}{Yes}, 
        \textbf{SkySenseGPT}: \textcolor{green}{Yes},  
        \textbf{VHM}: \textcolor{red}{no}.\\[4PT]
        \textit{What is the orientation of the road in the image?}\\
        \textbf{GT}: \textcolor{green}{North-South}, 
        \textbf{GeoChat}: \textcolor{red}{Right}, 
        \textbf{SkySenseGPT}: \textcolor{red}{horizontal},  
        \textbf{VHM}: \textcolor{green}{north-south}.\\[4PT]
    \end{minipage}
    
    \vspace{15pt}
  \begin{minipage}[t]{0.18\textwidth}
    \vspace{0pt}
    \includegraphics[width=\linewidth]{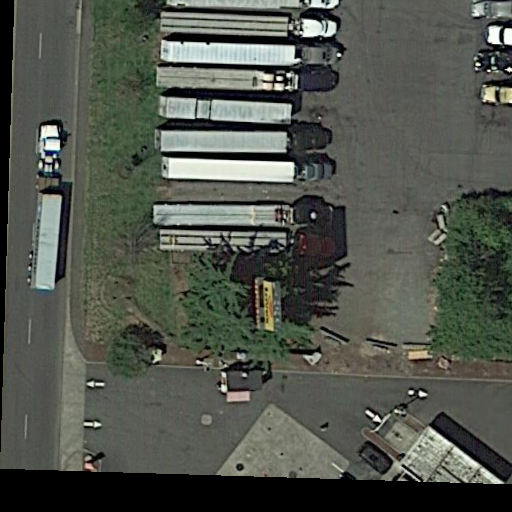}
    \smallskip
    \includegraphics[width=\linewidth]{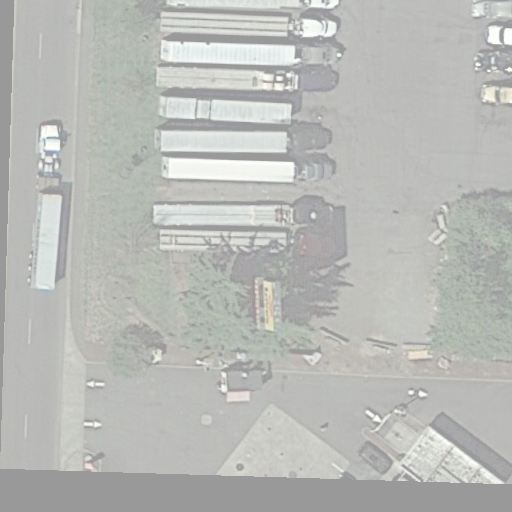}
    \smallskip
    \centering
    (b)
  \end{minipage}%
    \hspace{0.02\textwidth}
    \begin{minipage}[t]{0.77\textwidth}
        \vspace{0pt}
        \raggedright
        \small
        Top image: \textbf{P0007\_0004.png}\\
        Bottom image: \textbf{Corruption type:} Haze, \textbf{Severity:} 5.\\[6pt]
        \textit{What is the predominant color of the large vehicles?}\\
        \textbf{GT}: \textcolor{green}{white}, 
        \textbf{GeoChat}: \textcolor{red}{teal}, 
        \textbf{SkySenseGPT}: \textcolor{green}{white}, 
        \textbf{VHM}: \textcolor{green}{white}.\\[4PT]
        \textit{How are the large vehicles predominantly arranged in the parking lot?}\\
        \textbf{GT}: \textcolor{green}{in rows}, 
        \textbf{GeoChat}: \textcolor{red}{Cluster}, 
        \textbf{SkySenseGPT}: \textcolor{green}{in rows}, 
        \textbf{VHM}: \textcolor{red}{parked}. \\[4PT]
        \textit{Are there any large vehicles that are positioned away from the main cluster?}\\
        \textbf{GT}: \textcolor{green}{Yes}, 
        \textbf{GeoChat}: \textcolor{green}{Yes}, 
        \textbf{SkySenseGPT}: \textcolor{green}{Yes}, 
        \textbf{VHM}: \textcolor{green}{Yes}.\\[4PT]
        \textit{What is the orientation of the large vehicles in the parking lot?}\\
        \textbf{GT}: \textcolor{green}{north-south}, 
        \textbf{GeoChat}: \textcolor{red}{Vertical}, 
        \textbf{SkySenseGPT}: \textcolor{green}{parallel}, 
        \textbf{VHM}: \textcolor{green}{parallel}. \\[4PT]
    \end{minipage}
    
    \caption{Selected VQA examples. Correct answers are shown in green, and incorrect answers are shown in red.}
    \label{fig:vqa_visualise}

\end{figure}


\begin{figure*}[ht!]
    \centering
    \begin{subfigure}[t]{\textwidth}
        \centering
        \begin{subfigure}[t]{0.15\linewidth}
            \centering
            \includegraphics[width=\linewidth]{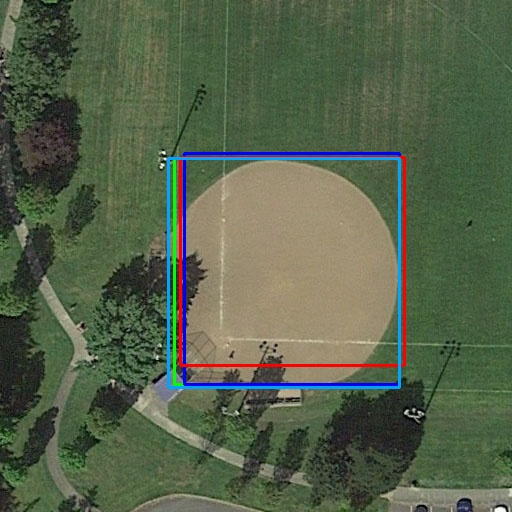}
        \end{subfigure}
        \begin{subfigure}[t]{0.15\linewidth}
            \centering
            \includegraphics[width=\linewidth]{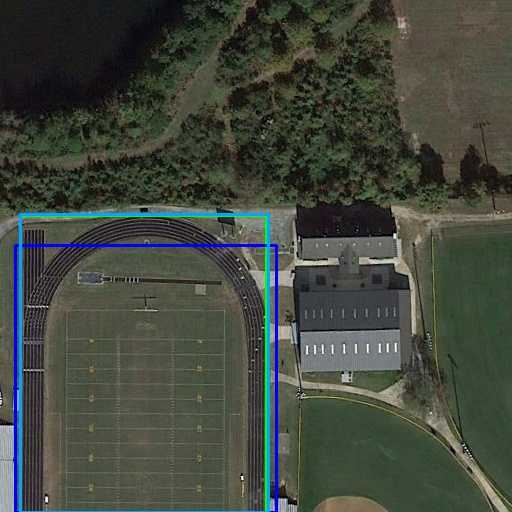}
        \end{subfigure}
        \begin{subfigure}[t]{0.15\linewidth}
            \centering
            \includegraphics[width=\linewidth]{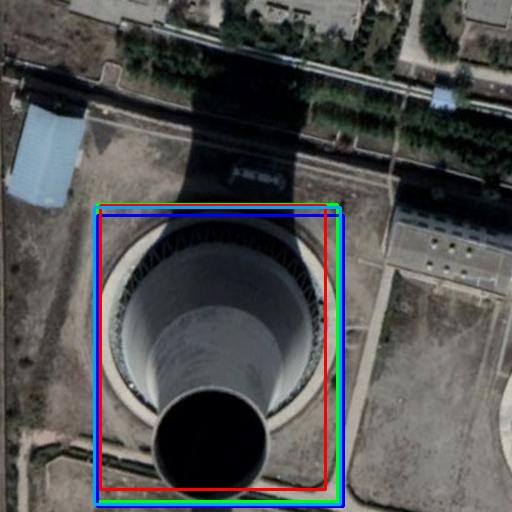}
        \end{subfigure}
        \begin{subfigure}[t]{0.15\linewidth}
            \centering
            \includegraphics[width=\linewidth]{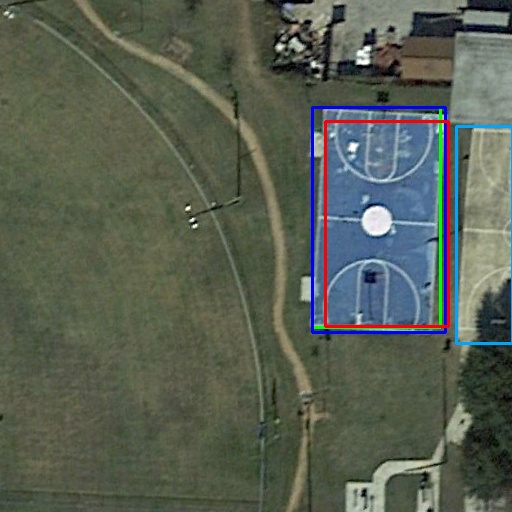}
        \end{subfigure}
        \begin{subfigure}[t]{0.15\linewidth}
            \centering
            \includegraphics[width=\linewidth]{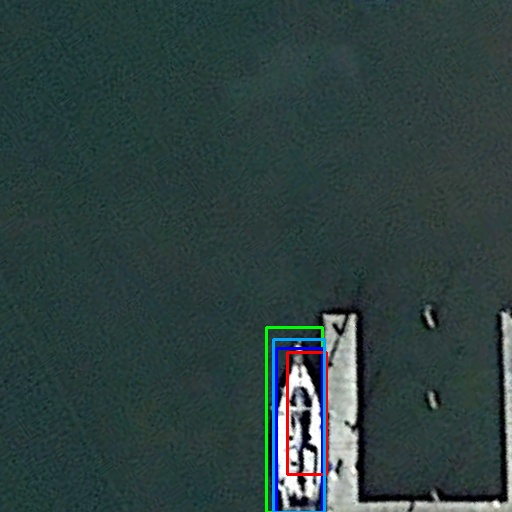}
        \end{subfigure}
        \begin{subfigure}[t]{0.15\linewidth}
            \centering
            \includegraphics[width=\linewidth]{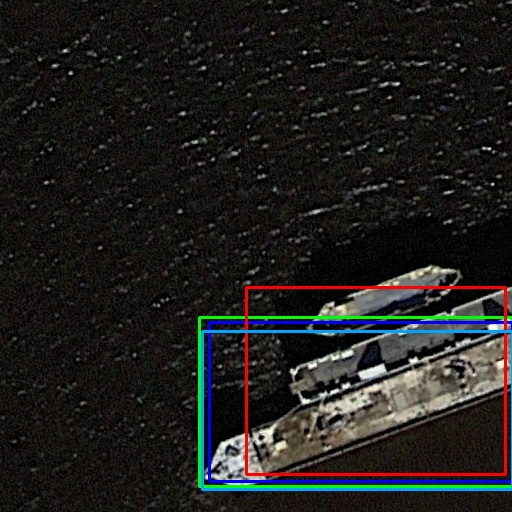}
        \end{subfigure}
    \end{subfigure}

    \vspace{10pt}
    
    \begin{subfigure}[t]{\textwidth}
        \centering
        \begin{subfigure}[t]{0.15\linewidth}
            \centering
            \includegraphics[width=\linewidth]{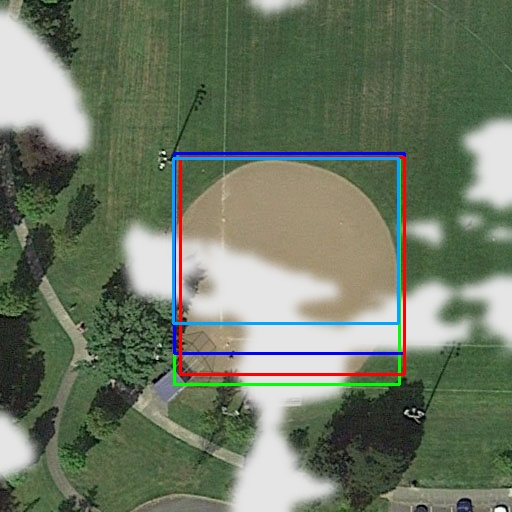}
        \end{subfigure}
        \begin{subfigure}[t]{0.15\linewidth}
            \centering
            \includegraphics[width=\linewidth]{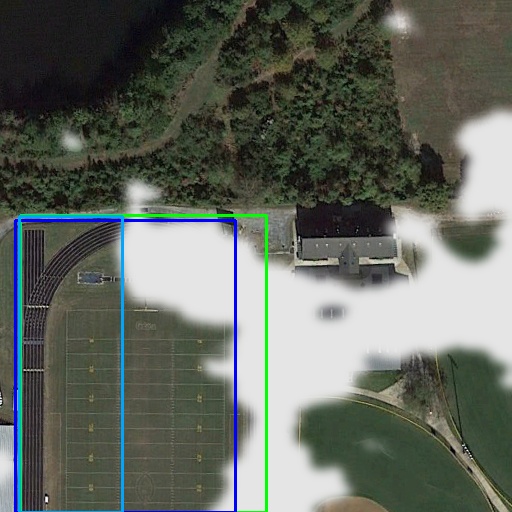}
        \end{subfigure}
        \begin{subfigure}[t]{0.15\linewidth}
            \centering
            \includegraphics[width=\linewidth]{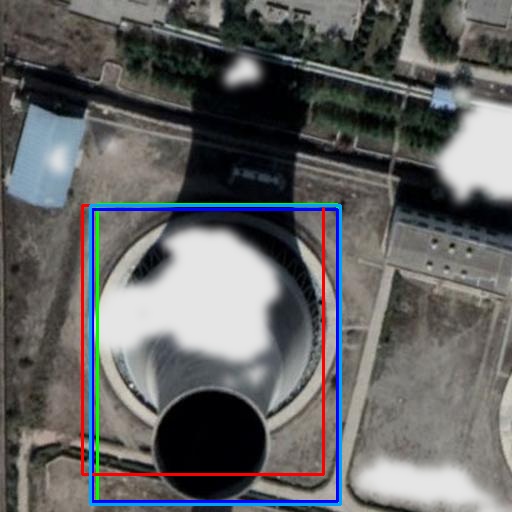}
        \end{subfigure}
        \begin{subfigure}[t]{0.15\linewidth}
            \centering
            \includegraphics[width=\linewidth]{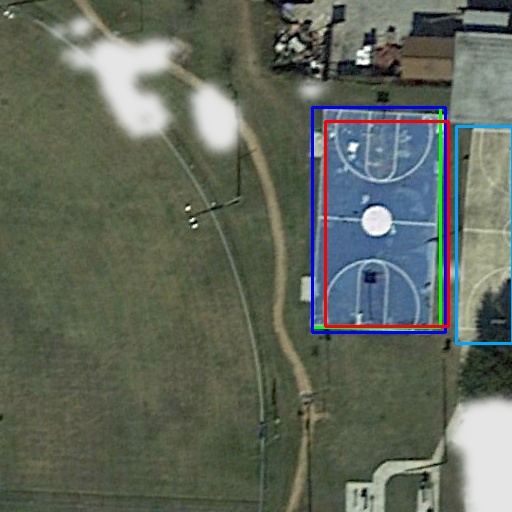}
        \end{subfigure}
        \begin{subfigure}[t]{0.15\linewidth}
            \centering
            \includegraphics[width=\linewidth]{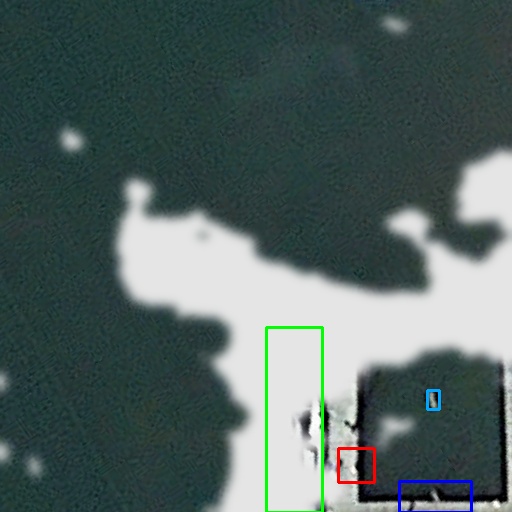}
        \end{subfigure}
        \begin{subfigure}[t]{0.15\linewidth}
            \centering
            \includegraphics[width=\linewidth]{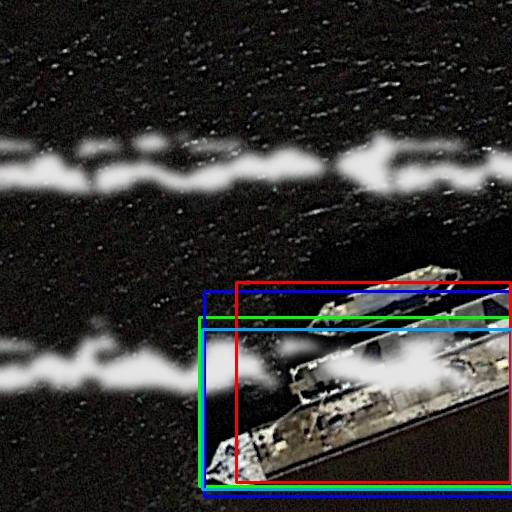}
        \end{subfigure}
    \end{subfigure}

    \vspace{10pt}

    \begin{subfigure}[t]{\textwidth}
        \centering
        \begin{subfigure}[t]{0.15\linewidth}
            \centering
            \includegraphics[width=\linewidth]{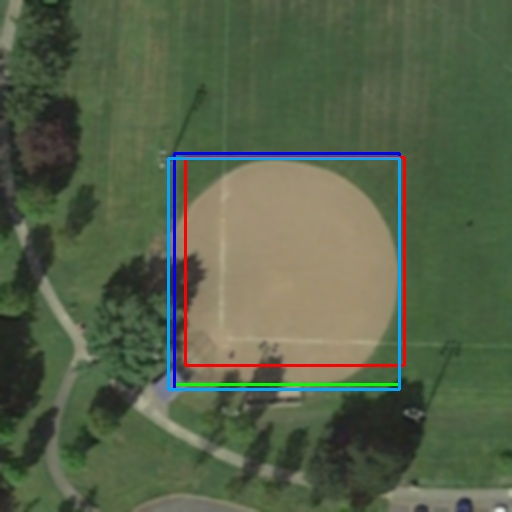}
        \end{subfigure}
        \begin{subfigure}[t]{0.15\linewidth}
            \centering
            \includegraphics[width=\linewidth]{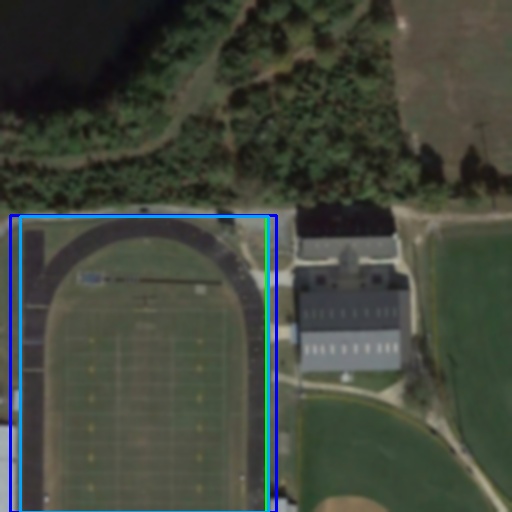}
        \end{subfigure}
        \begin{subfigure}[t]{0.15\linewidth}
            \centering
            \includegraphics[width=\linewidth]{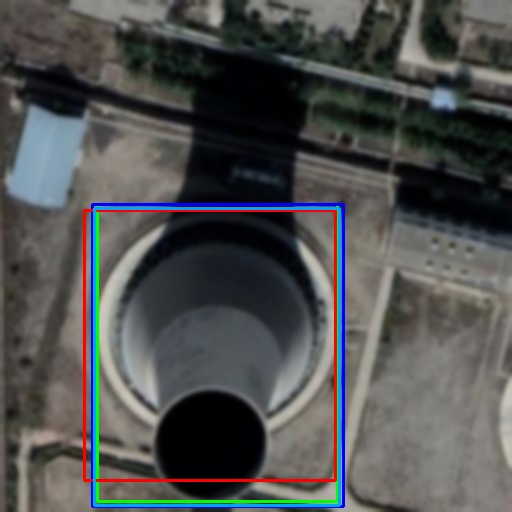}
        \end{subfigure}
        \begin{subfigure}[t]{0.15\linewidth}
            \centering
            \includegraphics[width=\linewidth]{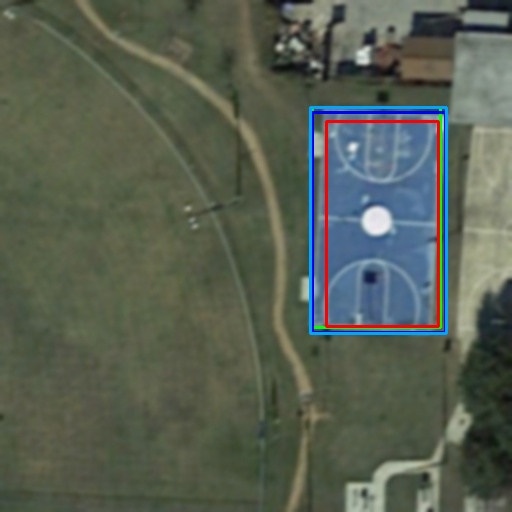}
        \end{subfigure}
        \begin{subfigure}[t]{0.15\linewidth}
            \centering
            \includegraphics[width=\linewidth]{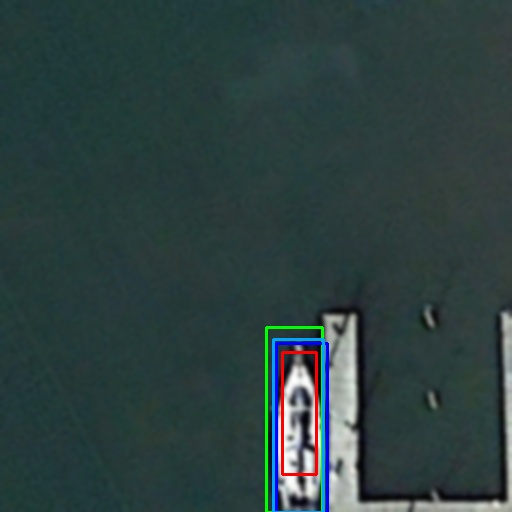}
        \end{subfigure}
        \begin{subfigure}[t]{0.15\linewidth}
            \centering
            \includegraphics[width=\linewidth]{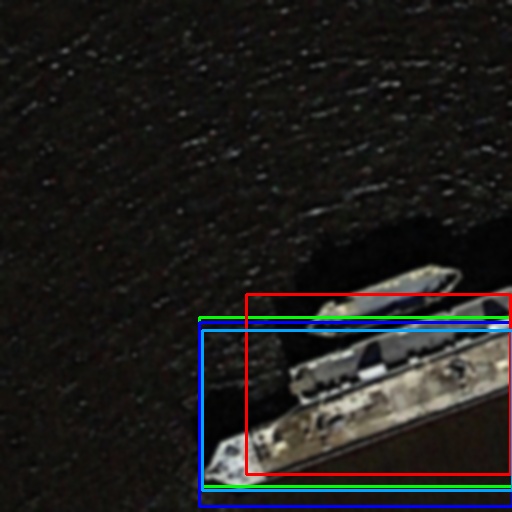}
        \end{subfigure}
        \caption*{\raisebox{0.7ex}{\fcolorbox{green}{white}{\rule{1ex}{0ex}}} GT\quad
                \raisebox{0.7ex}{\fcolorbox{GeoGround}{white}{\rule{1ex}{0ex}}} GeoGround\quad
                \raisebox{0.7ex}{\fcolorbox{VHM}{white}{\rule{1ex}{0ex}}} VHM\quad
                \raisebox{0.7ex}{\fcolorbox{Falcon}{white}{\rule{1ex}{0ex}}} Falcon\quad
                }
    \end{subfigure}

    \caption{Selected visual grounding examples. The top row presents results on the clean images, while the second and third rows illustrate results under clouds and Gaussian blur corruptions, respectively, both applied with severity level 5. From left to right, the visual grounding questions are: \textit{``The baseball field has a brown infield and is situated in the center of the image surrounded by green grass'', ``The ground track and field with multiple lanes is located next to a large building with a dark roof'', ``The chimney seen in the center of the image is cylindrical and extends from the middle to the bottom of the frame'', ``The basketball court featured in the image is colored in blue with white markings and is located on the right-hand side of the frame'', ``The small ship is situated vertically near the bottom right corner of the image'', and ``The ship positioned towards the bottom edge of the image''.}}
    \label{fig:grd_visualise}
\end{figure*}


\clearpage

\end{document}